\newif\ifAnonymous
\Anonymousfalse

\ifAnonymous
\documentclass[acmsmall,screen,review,anonymous,nonacm]{acmart}
\else
\documentclass[acmsmall,screen,nonacm]{acmart}
\setcopyright{none}
\fi

\ifAnonymous
\newcommand{\theauthor}{the author(s)}
\newcommand{\theauthorpos}{the author(s)'}
\else
\newcommand{\theauthor}{the author}
\newcommand{\theauthorpos}{the author's}
\fi

\usepackage{fancyvrb} % sophisticated verbatim text
	\fvset{tabsize=4}
\usepackage{amsmath} % standard maths package
\usepackage{bbm}
\usepackage{centernot} % horizontally-centered negation overlaid on symbols

\usepackage{booktabs}

\usepackage{enumitem} % For customizing enumeration labels

\usepackage{bussproofs} % sequent calculus
\usepackage{xcolor} % create red-coloured text for notes
	\newenvironment{notes}{\color{red}\noindent}{}
\usepackage{listings} % code listings that escape to math mode on '$'
	\lstdefinelanguage{diff}{% unified-diff colouring (first-column markers)
	  morecomment=[f][\color{gray}]{---},
	  morecomment=[f][\color{gray}]{+++},
	  morecomment=[f][\color{blue!70!black}]{@@},
	  morecomment=[f][\color{red!70!black}]-,
	  morecomment=[f][\color{green!45!black}]+,
	}
\usepackage{algorithm}
\usepackage[noend]{algpseudocode}
\makeatletter
	\def\BState{\State\hskip-\ALG@thistlm}
	\algnewcommand{\LineComment}[1]{\Statex \hskip\ALG@thistlm #1}
	\algnewcommand{\IndentLineComment}[1]{\Statex \hskip\ALG@tlm #1}
\makeatother
\usepackage{float} % flexible floating containers

\renewcommand{\epsilon}{\varepsilon}

\newcommand{\xnat}{x_{\mathit{nat}}}
\newcommand{\xtie}{x_{\mathit{t}}}
\newcommand{\xbinit}{x_{\mathit{init}}}
\newcommand{\Funname}[1]{\mathit{#1}}
\newcommand{\Truncate}{\Funname{Truncate}}

\newcommand{\ArgMax}{\Funname{ArgMax}}

\newcommand{\SqrtUB}{\Funname{SqrtUB}}
\newcommand{\FrobUB}{\Funname{FrobUB}}

\newcommand{\margin}[3]{m_{#2,#1}(#3)}
\newcommand{\MarginLipschitz}[2]{L_{#2,#1}}
\newcommand{\cat}{\mathbin{+\!\!+}}
\newcommand{\amul}{a_{\mathrm{mul}}}
\newcommand{\fl}[1]{\mathrm{fl}( #1 )}
\newcommand{\etamv}{\eta_{\mathrm{mv}}}
\newcommand{\defeq}{\triangleq}  % ``is defined as''
\newcommand{\R}{\mathbb{R}}
\newcommand{\adot}[1]{a_{\mathrm{dot}}(#1)}
\newcommand{\adotfwd}[1]{a_{\mathrm{dot}}^{\mathrm{fwd}}(#1)}
\newcommand{\asnorm}[1]{\|\,|#1|\,\|_2}  % spectral norm of entrywise absolute values
\newif\ifcomments
\commentsfalse

\usepackage{tikz}
\usetikzlibrary{shapes.geometric, arrows}

\usepackage{graphicx} % necessary for inserting .pdfs
\usepackage{color} % color used by hyperref which is needed by artifact badge
\definecolor{lcol}{rgb}{0,0,0.5}

\usepackage[capitalize]{cleveref}
\crefformat{equation}{(#2#1#3)}
\ifAnonymous
\crefname{appendix}{Supplemental Appendix}{Supplemental Appendices}
\Crefname{appendix}{Supplemental Appendix}{Supplemental Appendices}
\crefname{suppfigure}{Supplemental Fig.}{Supplemental Figs.}
\Crefname{suppfigure}{Supplemental Fig.}{Supplemental Figs.}
\else
\crefname{appendix}{Appendix}{Appendices}
\Crefname{appendix}{Appendix}{Appendices}
\crefname{suppfigure}{Fig.}{Figs.}
\Crefname{suppfigure}{Fig.}{Figs.}
\fi

\begin{document}

\title{Lipschitz-Based Robustness Certification Under Floating-Point Execution}
% extended version title is too long so we need to use this macro
%\titlerunning{Sensitivity-Based Robustness Certification}

\author{Toby~Murray}
\email{toby.murray@unimelb.edu.au}
\orcid{0000-0002-8271-0289}
\affiliation{%
  \institution{University of Melbourne}
  \city{Melbourne}
  \country{Australia}
}

%\inst{1}\orcidID{0000-0002-8271-0289}
%
%\authorrunning{T.~Murray}

%\institute{University of Melbourne, Australia}

\begin{abstract}
Lipschitz-based certification checks a neural network's robustness with a small numerical computation. The check takes milliseconds per input, and its implementation is small enough to be formally verified. As with most robustness verification, its soundness is proved against a network that executes in real arithmetic, whereas deployed networks execute in floating point. Floating-point rounding must be accounted for in two places: in the network's execution and, because the check is itself numerical, in the certifier's own computation.

We show that the conservatism of Lipschitz-based certification does not absorb the rounding in the network's execution: we exhibit counterexamples to the certificates of a formally verified real-arithmetic certifier that computes in exact arithmetic, and show that an adversary who controls the weights can make it issue false certificates for ordinary test images at standard perturbation bounds. We then present the first Lipschitz-based certification method that is both efficient and sound with respect to floating-point execution. The floating-point execution of a network is not a Lipschitz function, so the method keeps the real-arithmetic network's Lipschitz constant and then bounds separately, over the whole perturbation ball, how far the floating-point outputs can depart from the real-arithmetic ones, through a compositional theory of the deviation interleaved with sufficient conditions for the absence of overflow. For a model certified pre-deployment, we show how measurements of a high-precision execution tighten the bound. We compute the spectral norms on which all of this rests by Gram iteration in float64 with its rounding error tracked, and prove that it never under-estimates the true norm while running orders of magnitude faster than exact rational arithmetic.

We mechanise our theory in Rocq and implement it in a robustness certifier, which we evaluate on dense networks spanning image, tabular and many-class classification. It rejects every certificate a counterexample refutes and certifies entire test sets at milliseconds per instance on a CPU. The pre-deployment measurement reclaims many of the relatively few
real-arithmetic certificates that its floating-point soundness costs.
We compare it with ERAN, to our knowledge the only other public floating-point-sound verifier for networks of our size. At float32 ERAN's GPUPoly loses most of its certificates on most of our models to its own rounding accounting; ours does not, and is more than an order of magnitude faster.
\end{abstract}

\maketitle

\section{Introduction}\label{sec:introduction}\label{sec:intro}

Robustness is an important property for helping to ensure the trustworthiness
of neural network classifier outputs. Given a neural network~$N$ produces
output class $\ArgMax(N(x))$ for input~$x$, we say that this answer is \emph{robust}
if the neural network would have output the same class for any nearby input~$x'$ within distance~$\varepsilon$ of~$x$:
$\forall x'.\ \|x - x'\| \leq \varepsilon \Longrightarrow\ArgMax(N(x')) = \ArgMax(N(x))$.

Robustness is checked by two broad approaches. Robustness \emph{verifiers} reason symbolically about
the network, by abstract interpretation~\cite{DeepZ,DeepPoly} or specialised
solvers~\cite{reluplex}. Lipschitz-based \emph{certification}~\cite{leino2021,weng2018towards}
instead bounds the network's sensitivity by a Lipschitz constant and certifies an input~$x$ by
checking that the largest entry of~$N(x)$ exceeds every other entry by more than the constant
times~$\varepsilon$. The check is
thus a small numerical computation, which makes its implementation amenable to formal
verification~\cite{tobler2025} and takes milliseconds per input, so that certification is
applied to entire test sets and, in the most recent work, to billion-parameter
models~\cite{hu2026lipnext}.

Both approaches, with few exceptions~\cite{DeepZ,DeepPoly,FMIPVerify,song2021qnnverifier,DeepJ},
assume that the network executes in real arithmetic, whereas deployed networks execute in
floating point. This semantic gap between the arithmetic that the tool assumes and how networks execute is
an important PL challenge in neural network verification~\cite{cordeiro2025esop}, and it is known to produce
\emph{false certificates}: inputs that a verifier verifies robust although the network classifies
a nearby input differently~\cite{Jia_Rinard_SAS21}. Rounding bears on certification in a second
place as well, the certifier's own computation, which rational
arithmetic~\cite{tobler2025} or rounded interval arithmetic~\cite{pyrat2024,jin2024around} keeps
sound. Floating-point rounding must therefore be accounted for in two places: the network's
execution and the certifier's own computation.

We investigate the problem of rounding in both places in the context of Lipschitz-based
(global) robustness certification (\cref{sec:lipschitz}), where floating-point rounding has
received less attention. We present the first Lipschitz-based
certification check that is sound for the network's floating-point execution, under standard
floating-point semantics (\cref{sec:float-error}), while preserving its benefits mentioned above. That such a
check exists at all is not obvious: the floating-point execution of a network is not a Lipschitz
function, because rounding can move its output by a fixed amount however close two inputs are.
Our check therefore does not seek a Lipschitz constant for the floating-point execution. It keeps
the real-arithmetic network's constant and bounds, separately, how far the floating-point outputs
can depart from the real-arithmetic ones, not at the certified input alone but at every input
within~$\varepsilon$ of it. We begin (\cref{sec:cex}) by showing that the conservatism of
Lipschitz-based certification does not on its own rule out false certificates, even from a
formally verified real-arithmetic certifier, and that an adversary who controls the model
weights can obtain false certificates for ordinary test images at standard perturbation bounds.

Our theory covers feed-forward networks whose activations are Lipschitz and computed exactly in
floating point (\cref{sec:nn-model}), under the standard model of round-to-nearest with gradual
underflow~\cite{higham2002} (\cref{sec:float-error}). That model holds only in the absence of
overflow, so we first derive sufficient conditions for its absence (\cref{sec:overflow}), then a
compositional theory of the deviation between real and floating-point execution
(\cref{sec:deviation}), and from it the floating-point certification check
(\cref{sec:robustness}). That check demands more of the input than the real-arithmetic one:
the deviation bound is the additional demand. It therefore certifies fewer inputs; how many
fewer is the cost of floating-point soundness, which \cref{sec:eval} measures. For a model certified pre-deployment, we show how measurements of a
high-precision (float64) execution of it soundly tighten the deviation bound, reclaiming much of
that cost (\cref{sec:hybrid}). Both the check and the deviation bounds rest on the spectral
norms of the weight matrices, which Tobler et al. bound by Gram iteration~\cite{delattre2023} in
exact rational arithmetic, prohibitively slowly at scale. We show how the same iteration can be
run in float64 without sacrificing soundness, by tracking its rounding error and proving that the
result never under-estimates the true norm, and find it orders of magnitude faster
(\cref{sec:norms}).

We implement our approach in a robustness certifier for dense networks with ReLU hidden layers,
extending a re-implementation of Tobler et al.'s exact-rational certifier~\cite{tobler2025}, and
evaluate it (\cref{sec:eval}) on their globally robust~\cite{leino2021} image models~\cite{tobler2025} (MNIST, Fashion MNIST, CIFAR-10),
a tabular task (HIGGS), whose entire $500{,}000$-instance test set we certify, and a many-class
image task (EMNIST). Our certifier rejects every input whose certificate \cref{sec:cex} refutes, and
floating-point soundness costs little verified robust accuracy (the fraction of test points both
certified robust and correctly classified) on all models but CIFAR-10. The CIFAR-10 cost is governed by the certification
task rather than by the scale of the network and pre-deployment measurement reclaims most of it. To place these
results in context, we compare against ERAN~\cite{ERAN}, to our
knowledge the only publicly available floating-point-sound verifier demonstrated on networks of
our size,  on the five image models. At float32, our deployment format, ERAN's GPUPoly as shipped is
near-vacuous on four of the five models, and even with its float32
rounding under-charged (\cref{sec:rq3}) it loses most of its real-arithmetic certificates on
three models, whereas ours does not.
Ours verifies more on four of the five and almost ties
on the fifth, while being more than an order of magnitude faster.

We formalise all of our results in the Rocq theorem prover~\cite{Coq} (version 8.20.1) on top of
the LAProof library~\cite{LAProof}, a formalisation of standard floating-point linear-algebra
error models~\cite{higham2002}\ifAnonymous; the proof scripts are provided as supplementary
material\fi. The only axioms our Rocq development introduces concern standard properties of
matrix spectral norms, listed in full in \cref{app:axioms}. The executable certifier we use in
our evaluation is a separate, unverified Python implementation; \cref{sec:eval-setup} states the
resulting trust boundary.

\paragraph{Use of generative AI}
Generative AI (Claude Code and ChatGPT) was used throughout this research, under \theauthorpos{}
direction, with \theauthor{} reviewing and guiding each step in addition to auditing final outputs.
The key ideas were developed in iterative discussion between \theauthor{} and generative AI,
first worked out as informal pen-and-paper mathematics before being formalised in Rocq; the
Rocq formalisation was likewise developed with generative AI, with \theauthor{} auditing its
definitions, axioms (\cref{app:axioms}), and top-level theorem statements for faithfulness to the
intended mathematics. Generative AI was also used to implement the Python certifier (\cref{sec:eval-setup} explains the
trust boundary and how the certifier's implementation was validated) and the surrounding experimental
infrastructure (model training, benchmarking, and analysis), and to develop parts
of the counterexample search (\cref{sec:cex}) and the modification of GPUPoly that
\cref{app:gpupoly} describes. Generative AI was also used in drafting and revising this
manuscript. The author\ifAnonymous(s) take\else{} takes\fi{} full responsibility for the paper's
contents.

\section{Background and Motivation}\label{sec:background}\label{sec:motivation}

\subsection{Robustness and Lipschitz-Based Certification}\label{sec:lipschitz}

We view a neural network as a function~$N$ from input vectors~$x$ to output vectors~$y$ of
fixed dimensions, and focus exclusively on classifiers, where the output class of~$y$ is the
index of its maximum component, $\ArgMax(y)$, among $|y|$ classes. We consider $\ell_2$
robustness, with distance measured by the Euclidean norm (written $\|v\|_2$ for vector~$v$):

\begin{definition}\label{defn:robust}
  Given $\varepsilon \ge 0$, input~$x$, and output~$y = N(x)$, we say that $x$
  is \emph{robust} iff:
  \[
  \forall x'. \ \|x - x'\|_2 \leq \varepsilon \Longrightarrow \ArgMax(N(x')) = \ArgMax(y) .
  \]
\end{definition}
We write $B(x,\varepsilon) \defeq \{x' \mid \|x - x'\|_2 \le \varepsilon\}$ for the closed $\varepsilon$-ball centred at~$x$.
A certifier is \emph{sound} if it never certifies a non-robust input as robust, and \emph{vacuous} if it
certifies no input at all.

Fix an input~$x$ and its associated output $y = N(x)$.
Let $i^* = \ArgMax(y)$ denote the output class of~$x$, and let~$j$ be any other \emph{competing} class.
We write $\margin{i^*}{j}{x}$ to denote the \emph{margin} between the output and competing class at input~$x$, which is simply
$y_{i^*} - y_j$. This definition generalises to arbitrary inputs~$x'$:
$\margin{i^*}{j}{x'} = N(x')_{i^*} - N(x')_j$.
Robustness of~$x$ follows whenever, for each competing class $j \not= i^*$, the margin
$\margin{i^*}{j}{x'}$ remains positive for all $x'$ within $\varepsilon$ of~$x$.

Leino et al.~\cite{leino2021} check margin positivity by computing for each~$j$ a
margin Lipschitz constant $\MarginLipschitz{i^*}{j}$, which bounds the degree to which the margin $\margin{i^*}{j}{\cdot}$ changes per change in
the input:\[
  \forall x_1, x_2. \ \frac{|\margin{i^*}{j}{x_1}-\margin{i^*}{j}{x_2}|}{\|x_1-x_2\|_2}\le \MarginLipschitz{i^*}{j} .
\]

Margin Lipschitz constants are derived from the spectral norms of the layer weight matrices.
These spectral norms cannot be computed exactly; they are approximated iteratively, either from
below (as in power iteration~\cite{gouk2021}) or from above (as in Gram
iteration~\cite{delattre2023}). Since sound certification requires an \emph{upper} bound on the
network's sensitivity, the from-above approach is most appropriate; we return to computing
these bounds soundly and at scale in \cref{sec:norms}.
With margin Lipschitz constants computed, certifying robustness for an input~$x$  requires simply checking for each~$j \not= i^*$:
\begin{equation}\label{eqn:certification-check}
\margin{i^*}{j}{x} > \MarginLipschitz{i^*}{j} \cdot \varepsilon .
\end{equation}

\subsection{Floating-Point Counterexamples to Real-Arithmetic Certificates}\label{sec:cex}

Tobler et al.'s formally verified certifier~\cite{tobler2025} computes the margin Lipschitz
constants and the check~\cref{eqn:certification-check} in arbitrary-precision rational
arithmetic, so that no rounding occurs in the certifier's own computation. But, as Tobler et al.
note, its answers remain sound
only if the network itself executes in real arithmetic, whereas deployed networks execute in
floating point. This raises the question of whether rounding in the network's execution can make
the certifier issue a \emph{false certificate}: certify an input~$x$ robust, as it would be under
real arithmetic, when the network's floating-point execution classifies some~$x'$
within~$\varepsilon$ of~$x$ differently.

Rounding is known to produce false certificates from \emph{complete} robustness
verifiers~\cite{Jia_Rinard_SAS21,zombori2021fooling}, whose analysis is tight in the sense that
it never under-estimates robustness. A Lipschitz-based certifier is not complete: to be sound and
efficient, it over-estimates the network's sensitivity, through a \emph{global} margin Lipschitz
constant, and therefore under-estimates robustness. For this reason it is not obvious that such
a certifier should be vulnerable in the same way: perhaps the conservatism that this
over-estimation introduces is sufficient to absorb the rounding.

We asked this question of Tobler et al.'s certifier, on the three float32 models on which it was
originally evaluated, namely MNIST~\cite{MNIST}, Fashion MNIST~\cite{Fashion_MNIST}, and
CIFAR-10~\cite{CIFAR10}, whose robustness Tobler et al. evaluated at $\varepsilon = 0.3$, $0.25$ and
$0.141$ respectively (\cref{sec:eval}). A \emph{counterexample} is a pair of inputs $x$ and $x'$
with $\|x' - x\|_2 \le \varepsilon$ that the network's floating-point execution classifies
differently, although the certifier certifies $x$ robust at~$\varepsilon$. Throughout, that
execution is in float32 unless stated otherwise, and is IEEE-754-compliant (every operation
individually rounded to nearest, with gradual underflow), and every pair we report is
re-verified in exact arithmetic, independently of the search that found it
(\cref{app:cex-figs}).

We first asked whether any counterexample exists at all. A simple search (\cref{app:cex-figs}),
which crosses the decision boundary from a test image~$\xnat$ with the DeepFool adversarial
attack~\cite{DeepFool} and bisects along the line back towards~$\xnat$, finds counterexamples
for each of the three models; we stopped after collecting $30$ per model. The radii~$\varepsilon$ are
minuscule---from roughly $4\times10^{-8}$ to $2\times10^{-6}$ across the three models---many
orders of magnitude below any perturbation of interest, and the certified points~$x$ remain
recognisable images (\cref{fig:cex_float32}). The radii track the format's rounding: against
the same models executed at float16 the search finds counterexamples at radii from roughly
$4\times10^{-4}$ to $9\times10^{-3}$, and at float64 from $10^{-16}$ to $5\times10^{-15}$. These counterexamples establish that the
certifier's conservatism does not make its certificates sound.

We next asked the question as a user of the certifier would: at the model's standard
radius~$\varepsilon$, is any certified test image refutable by another image? Here $x$ is an
image~$\xnat$ drawn from the model's test set rather than a point derived from one, and ``image'' means that the pixel
values (per colour channel) are 8-bit intensities $k/255$ with $k$ an integer, so that both $x$
and $x'$ lie on the 8-bit \emph{input grid}. We refer to this search as the
\emph{grid search}, and to the earlier one, whose points range over all float32 inputs, as the
\emph{continuous search}. For each test image~$x$ that the certifier certifies at the
standard~$\varepsilon$ from the model's float32 output for~$x$, a projected-gradient (PGD)
attack~\cite{madry2018} searches the grid points of the $\varepsilon$-ball around~$x$ for an
image that the float32 execution classifies differently from~$x$ (\cref{app:cex-figs}).

On the three naturally trained models this search finds nothing: no certified test image has
an image within the standard~$\varepsilon$ that the float32 execution classifies differently.
A certified image~$x$ can be refuted by an
image~$x'$ within~$\varepsilon$ only if the rounding errors in the two floating-point
margins, for~$x$ and for~$x'$, are together large enough: they must exceed the slack by
which~$x$ passed the certification check plus the conservatism of the check itself, the amount
by which $\MarginLipschitz{i^*}{j}\,\varepsilon$ over-estimates how far the real-arithmetic
margin falls between~$x$ and~$x'$. For naturally trained float32 models like these, the
rounding errors are very small, while most test images are certified with slack to spare. In
this practical sense a real-arithmetic certifier is sound for these models, inputs and
perturbation bounds, but only because the rounding errors are tiny, a property of the model's
arithmetic that the certifier never examines.

\begin{figure}[t]
  \centering\includegraphics[width=\textwidth]{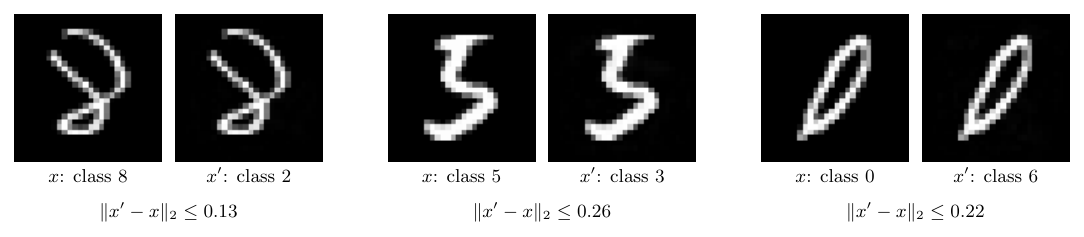}
  \caption{Counterexamples for the adversarially-biased float32 MNIST model, which retains
    $98.33\%$ test accuracy.\label{fig:cex_mnist_biased} In each pair, the test image $x$
    (left) is certified robust at $\varepsilon = 0.3$ by Tobler et al.'s verified
    real-arithmetic certifier, yet the image $x'$ (right) within $\varepsilon$ of $x$ is
    classified differently under float32 execution.}
\end{figure}

Our final experiment makes this point directly, under a stronger threat model in which the
adversary controls the model weights. Starting from the MNIST model, whose layers, like
those of all of Tobler et al.'s models, carry no bias terms, we add a large constant bias
($B = 10^6$) to every neuron of the last hidden layer, and give the output layer the
compensating bias that cancels the constant's contribution through its weight matrix. In real
arithmetic the two cancel (except for the ReLU's small effect on which neurons are active), and because biases do not
affect the Lipschitz constants, the real-arithmetic certification is entirely unchanged. Under
floating-point arithmetic, however, the inflated values leaving the last hidden layer make the
output layer's matrix-vector product accumulate rounding errors in proportion to them;
subtracting the compensating bias then recovers the small true signal but leaves those errors
behind, now enormous relative to the signal. This amplifies the deviation between the floating-point and real-arithmetic network
outputs by about six orders of magnitude, while the slack and conservatism of every certificate
are essentially unchanged. The adversarial model retains $98.33\%$ test accuracy (compared to
$98.40\%$ for the original), and the certifier, whose real-arithmetic view of the model is
unchanged, still certifies $9{,}549$ of the $10{,}000$ test images at $\varepsilon = 0.3$.
For $24$ of those certified images ($0.25\%$ of them) the grid search finds an image within
$\varepsilon$ that the float32 execution classifies differently (\cref{fig:cex_mnist_biased}
shows three). The analogous models for Fashion MNIST and CIFAR-10 (with $B = 3\times10^6$ and
$4\times10^6$ respectively, each losing under two percentage points (pp) of test accuracy) behave
the same way, and worse: $204$ ($2.5\%$) and $370$ ($7.9\%$) refuted certificates respectively.
The search is incomplete, so these counts are lower bounds. Both images of every pair have 8-bit pixel
values, so restricting a certifier to such inputs would not help (and inputs such as HIGGS's,
\cref{sec:eval}, lie on no grid at all).

The two cases differ only in the size of the model's rounding errors: tiny in the naturally
trained models, where the false certificates exist only at radii of no practical interest, and
large in the biased ones, where ordinary test images are falsely certified at the standard
radii, in models of ordinary accuracy, from the same Lipschitz constants and near-identical
margins. A certifier that reasons in
real arithmetic cannot tell the two cases apart, and no larger Lipschitz constant can absorb
the rounding errors, since the floating-point execution is not a Lipschitz function
(\cref{sec:intro}).
Making certification sound therefore requires a sound bound on how far rounding takes the
model's outputs from their real-arithmetic values, which the rest of this paper develops
(\cref{sec:float-semantics,sec:robustness}); what that bound costs on the models one would
deploy is one of the questions that \cref{sec:eval} answers.

\section{Floating-Point Semantics and Deviation from Real Arithmetic}\label{sec:float-semantics}

This section formalises, for dense feed-forward neural networks, the discrepancy between
real-arithmetic reasoning and floating-point execution behind the counterexamples of
\cref{sec:cex}.

\subsection{Feed-Forward Neural Networks}\label{sec:nn-model}

Over real arithmetic, such a network is composed of $L$ \emph{layers}, and layer~$\ell \in \{1,\ldots,L\}$
comprises a weight matrix~$W_{\ell} \in \mathbb{R}^{m_\ell \times n_\ell}$, a bias vector~$b_\ell \in \mathbb{R}^{m_\ell}$ and an
activation function~$\varphi_\ell$; it maps the $n_\ell$-element output of layer~$\ell-1$ to an $m_\ell$-element
output. For an input~$x \in \mathbb{R}^{n_1}$, write $z_{\ell-1}(x)$ for the input to layer~$\ell$,
with $z_0(x) = x$. Layer~$\ell$ computes the \emph{preactivation}~$a_\ell$ and then its output~$z_\ell$:
\begin{equation}\label{eqn:layer-semantics}
a_\ell = W_\ell z_{\ell-1} + b_\ell , \qquad z_\ell = \varphi_\ell(a_\ell) .
\end{equation}

\label{sec:activation-assumptions}
We make the following assumptions on activation functions throughout:
\begin{enumerate}[label=(\roman*)]
  \item \emph{Pointwise scalar maps}: each~$\varphi_\ell$ is a scalar function applied elementwise.
  \item \emph{Zero-preserving}: $\varphi_\ell(0) = 0$.
  \item \emph{Lipschitz continuous}: $\varphi_\ell$ has Lipschitz constant~$L(\varphi_\ell) \ge 0$.
  \item \label{assumption-fp-exact} \emph{FP-exact}: $\varphi_\ell(a)$'s floating-point evaluation doesn't induce rounding, for all finite~$a$.
\end{enumerate}
Assumptions~(ii) and~(iii) together give $\|\varphi_\ell(a)\|_2 \le L(\varphi_\ell)\,\|a\|_2$ for
every preactivation~$a$: the Lipschitz bound measured from~$0$, which~(ii) makes a fixed point, and
on which the radii of \cref{sec:overflow} rest.
Our results hold for any activations satisfying (i)--(iv), as ReLU, identity, absolute value
and hard tanh do. The networks we evaluate use ReLU (with $L(\varphi_\ell) = 1$) in hidden
layers and identity ($\varphi_L(z) = z$) at the output layer, and only our certifier
(\cref{sec:eval-setup}) is specific to ReLU and identity.
Assumption~\ref{assumption-fp-exact} excludes smooth activations such as sigmoid and tanh,
whose floating-point implementations introduce rounding; \cref{sec:conclusion} says what
admitting them would take.

\subsection{Floating-Point Error}\label{sec:float-error}

We adopt the standard model of floating-point error~\cite{higham2002}, in which each floating-point
operation is modelled in terms of its real arithmetic counterpart. It applies to the IEEE-754 formats
and is formalised by LAProof~\cite{LAProof}, on which our Rocq
formalisation rests. We assume round-to-nearest, ties-to-even with gradual underflow, the default in
floating-point implementations. Some runtimes instead \emph{flush} subnormal results to zero;
\cref{app:ftz} extends the theory to flush-to-zero semantics, where only the absolute-error constants
change (that extension is not mechanised).

The model has two positive constants, fixed by the format and the rounding mode. The \emph{unit
roundoff}~$u$ bounds the \emph{relative} error of a rounded operation; under round-to-nearest in
a format with $p$ significand bits, $u = 2^{-p}$.
The constant $\amul$ bounds the \emph{absolute} error of a multiplication or division whose result
is subnormal, where no relative bound holds; it is half the smallest positive subnormal of the
format. Its subscript records that, under gradual underflow, only multiplication and division incur
an absolute term: a sum or difference of floating-point numbers whose exact value is subnormal is
itself representable~\cite[\S2.12 and Problem~2.19]{higham2002}. For float16, float32 and float64,
$(p, u, \amul)$ is $(11, 2^{-11}, 2^{-25})$, $(24, 2^{-24}, 2^{-150})$ and $(53, 2^{-53}, 2^{-1075})$.
Directed rounding modes, whose relative error is at most $2u$, are covered by substituting $2u$
and $2\amul$ for $u$ and $\amul$ respectively, though LAProof covers only round-to-nearest, so
this extension is not mechanised.
We write $\fl{\cdot}$ for the floating-point counterpart of a real-valued computation.

\paragraph{Basic operations}
The following error model, which goes back to Demmel~\cite{demmel1984} and is Higham's
model~(2.8)~\cite[\S2.12]{higham2002}, applies to a basic operation~$\circ$ whose result does not overflow:\begin{equation}\label{eqn:mixed-model-basic-ops}
\fl{ x \circ y } = (x \circ y)(1 + \delta) + \eta, \qquad |\delta| \le u, \quad |\eta| \le \amul
\end{equation}
where $\delta$ and $\eta$ are \emph{input-dependent} terms that account respectively for the relative and
absolute error; for addition and subtraction, $\eta = 0$.
We write $F_{\max}$ for the largest finite value of the format. A result \emph{overflows} when it
rounds to a non-finite value. An exact result of magnitude below $F_{\max}$ cannot overflow, and our
overflow checks adopt this sufficient condition.

\paragraph{Matrix-vector product and bias addition}
The model for the matrix-vector product of $W \in \R^{m \times n}$ and $z \in \R^n$ is
LAProof's mixed error model for dot products~\cite{LAProof}, applied row by row:
\begin{equation}\label{eqn:mixed-model-matvec}
\fl{W z} = (W + \Delta{W})z + \etamv, \qquad |\Delta{W}| \le \gamma_n|W|, \quad \gamma_n \defeq (1 + u)^n - 1,
\end{equation}
where $|M|$ denotes the entrywise absolute value of a matrix or vector~$M$, $\le$ on matrices and
vectors is entrywise, and $\Delta W$ depends on $W$ and~$z$.
We use $\gamma_n$ in this exact form rather than the familiar closed form $\gamma_n \le nu/(1-nu)$:
the exact form is defined for all~$n$ and~$u$, so soundness does not require $nu < 1$ (what the
regime $nu \ge 1$ costs in tightness is the subject of \cref{sec:nell-cdot-u-lt-1}). The bound holds
whatever order the $n$ products are accumulated in~\cite[\S3.1, eq.~(3.4)]{higham2002}, so it covers
any dot-product implementation; our mechanisation fixes LAProof's sequential order.
The vector $\etamv\in\R^m$ collects the per-product absolute errors of each length-$n$ dot product
and satisfies $|\etamv| \le \adot{n}\,\mathbbm{1}$ with $\adot{n} \defeq (1+\gamma_n)\,n\,\amul$,
where $\mathbbm{1}$ is the all-ones vector in $\R^m$, so that $\|\etamv\|_2 \le \adot{n}\sqrt{m}$.
For the bias addition $\fl{v + b}$ of a preactivation~$v$ and bias~$b \in \R^m$, the error is purely
relative, since $\eta = 0$ for addition in \cref{eqn:mixed-model-basic-ops}:
$\fl{v + b} = v + b + \delta_{\mathrm{add}}$ with $|\delta_{\mathrm{add}}| \le u\,(|v| + |b|)$,
so that $\|\delta_{\mathrm{add}}\|_2 \le u\,(\|v\|_2 + \|b\|_2)$.

\paragraph{Floating-point network execution and deviation}
The \emph{floating-point evaluation} of the network replaces each
arithmetic operation with its floating-point counterpart. The \emph{deployment format} is the floating-point format in which the network executes once
deployed, whose constants $u$ and $\amul$ the error model above takes; it is float32 for every
model we evaluate. The weights $W_\ell$ and biases $b_\ell$ are finite values of that format (as
trained models' parameters are), and the network's inputs are the \emph{format-representable}
vectors: those whose components are finite values of the deployment format. A guarantee for every
format-representable input in a region therefore covers every input that the deployed network can
receive from that region.
Given floating-point activations $\hat{z}_{\ell-1}$ from the previous layer,
with $\hat{z}_0 = x$ (the input is used exactly, being format-representable), the
floating-point preactivation~$\hat{a}_\ell$ and activation~$\hat{z}_\ell$ at layer~$\ell$, and the
\emph{floating-point deviation}~$d_\ell$ of that activation from its real-arithmetic
counterpart, are
\begin{equation}\label{eqn:defn-d}
  \hat{a}_\ell \;\defeq\; \fl{\fl{W_\ell \hat{z}_{\ell-1}} + b_\ell}, \qquad
  \hat{z}_\ell \;\defeq\; \varphi_\ell(\hat{a}_\ell), \qquad
  d_\ell \;\defeq\; \hat{z}_\ell - z_\ell .
\end{equation}
Only the matrix-vector product and the bias addition round, not the activation
(assumption~\ref{assumption-fp-exact}). The deviation measures the semantic gap between real
arithmetic and floating-point execution.

\section{Sound Conditions for Absence of Overflow}\label{sec:overflow}

The error model of \cref{sec:float-error} applies only in the absence of overflow, so before
bounding the floating-point deviation (\cref{sec:deviation}) we must certify that none occurs,
and not just at~$x$ but uniformly over $B(x,\varepsilon)$. Writing $z_\ell(x')$ and
$\hat{z}_\ell(x')$ for the exact and floating-point activations at layer~$\ell$ on input~$x'$
(and likewise $d_\ell(x')$), this means bounding $\|\hat{z}_{\ell-1}(x')\|_2$ for every $x'$ in
the ball, which we do with two per-layer scalars computable from the weights and the format
constants alone: a \emph{radius} $r_{\ell-1} \ge \|z_{\ell-1}(x')\|_2$ and a \emph{deviation
bound} $D_{\ell-1} \ge \|d_{\ell-1}(x')\|_2$, each for all $x' \in B(x,\varepsilon)$.
The deviation bounds are derived in \cref{sec:deviation}, and their derivation relies on the
error model, which holds only once the layer producing the deviation is known not to overflow.
The two analyses are therefore \emph{interleaved}, proceeding forward through the network one
layer at a time: the overflow check at layer~$\ell$ consumes the deviation bound $D_{\ell-1}$
already established for layer~$\ell-1$, and the deviation bound $D_\ell$ then rests on that
check. Nothing is circular, because the base case needs no deviation bound at all: for
$\ell = 1$ the input is used exactly ($\hat{z}_0 = z_0 = x$), so $D_0 = 0$ and
$r_0 = \|x\|_2 + \varepsilon$. The radius alone does not suffice at later layers, since the
floating-point activation, which is what is actually fed into the computation, may differ from
the real one by up to $D_{\ell-1}$; by the triangle inequality,
$\|\hat{z}_{\ell-1}(x')\|_2 \le \|z_{\ell-1}(x')\|_2 + \|d_{\ell-1}(x')\|_2 \le r_{\ell-1} + D_{\ell-1}$
uniformly over $B(x,\varepsilon)$.

\paragraph{Deterministic radii}\label{sec:radii}
The \emph{deterministic radii} are defined inductively by
\[
  r_0 \;\defeq\; \|x\|_2 + \varepsilon, \qquad
  r_\ell \;\defeq\; L(\varphi_\ell)\bigl(\|W_\ell\|_2\, r_{\ell-1} + \|b_\ell\|_2\bigr),
\]
where $\|W_\ell\|_2$ is the spectral norm of~$W_\ell$, and $\|z_\ell(x')\|_2 \le r_\ell$ for all
$x' \in B(x,\varepsilon)$. The recurrence mirrors the layer semantics~\cref{eqn:layer-semantics},
each factor the worst case of one of its operations: the spectral norm is the most the
matrix-vector product can scale a vector, the bias adds at most its norm, and the activation-norm
bound of \cref{sec:nn-model} scales by at most~$L(\varphi_\ell)$.

\paragraph{Overflow quantities}
To certify that $\fl{W_\ell \hat{z}_{\ell-1}}$ does not overflow, each output entry, the dot product
of a row $w_{\ell,i}$ of~$W_\ell$ with~$\hat{z}_{\ell-1}$, must be finite. Two quantities govern
this. Rounding errors accumulate in proportion to the magnitudes of the individual
products regardless of their signs, so the first is the sum of absolute products
$\sum_j |w_{\ell,i,j}|\,|\hat{z}_{\ell-1,j}|$, which Cauchy--Schwarz bounds by
$\|w_{\ell,i}\|_2\,(r_{\ell-1} + D_{\ell-1})$. The second is the largest single product, which must
not itself overflow, bounded by $\|w_{\ell,i}\|_\infty\,(r_{\ell-1} + D_{\ell-1})$. Taking the worst
case over rows gives two scalars per layer,
\[
  S_\ell \;\defeq\; \max_i\,\|w_{\ell,i}\|_2 \cdot (r_{\ell-1} + D_{\ell-1}), \qquad
  M_\ell \;\defeq\; \max_i\,\|w_{\ell,i}\|_\infty \cdot (r_{\ell-1} + D_{\ell-1}),
\]
which depend only on the fixed weights, $r_{\ell-1}$ and~$D_{\ell-1}$.

Overflow of a length-$n$ dot product depends on these two quantities alone. Let
$\adotfwd{n} \defeq (1+\gamma_{n-1})\,n\,\amul$, with $\gamma_{n-1}$,
because the dot product has $n-1$ additions. Then we have:

\begin{lemma}[Forward error for FP dot products]\label[lemma]{lem:forward-error}
Let $a, b$ be length-$n$ vectors of finite floating-point values. If $\sum_j|a_j|\,|b_j| \le S$,
$|a_j|\,|b_j| \le M$ for all~$j$, $M < F_{\max}$, and
$S\,(1+\gamma_n) + \adotfwd{n} < F_{\max}$, then $\fl{a \cdot b}$ is finite and
$
  \bigl|\fl{a \cdot b}\bigr| \;\le\; S\,(1 + \gamma_n) + \adotfwd{n}.
$
\end{lemma}

Instantiated for every row with $S = S_\ell$, $M = M_\ell$ and $n = n_\ell$ (the input
dimension of layer~$\ell$, the full length of the dot products, zero weights included),
the lemma bounds every entry of $\fl{W_\ell \hat{z}_{\ell-1}}$ by
$S_\ell\,(1+\gamma_{n_\ell}) + \adotfwd{n_\ell}$. Adding the bias needs no rounding factor of its own:
if the exact sum of that bound and $\|b_\ell\|_\infty$ lies below the representable $F_{\max}$,
round-to-nearest returns a finite value.

\begin{theorem}[Layerwise overflow-freedom]\label{thm:overflow-free}
Let $\hat{z}_{\ell-1}$ be a finite floating-point vector with
$\|\hat{z}_{\ell-1}\|_2 \le r_{\ell-1} + D_{\ell-1}$, and let $W_\ell$, $b_\ell$ be finite
floating-point weights and bias. If
$S_\ell\,(1 + \gamma_{n_\ell}) + \adotfwd{n_\ell} + \|b_\ell\|_\infty < F_{\max}$
and $M_\ell < F_{\max}$, then the preactivation $\hat{a}_\ell = \fl{\fl{W_\ell \hat{z}_{\ell-1}} + b_\ell}$ is finite.
\end{theorem}

By assumption~\ref{assumption-fp-exact}, finiteness of the preactivation gives finiteness of
the activation $\hat{z}_\ell = \varphi_\ell(\hat{a}_\ell)$, so the theorem covers the whole layer.
Applied layer by layer, interleaved with the deviation bounds of \cref{sec:deviation}, it
reduces overflow certification over $B(x,\varepsilon)$ to two scalar inequalities per layer.

\paragraph{Practical applicability}\label{sec:nell-cdot-u-lt-1}
The analysis is sound at every width and precision, but only useful where
$n_\ell \cdot u \ll 1$ at every layer, since $\gamma_{n_\ell}$ grows roughly linearly in
$n_\ell \cdot u$ there and superlinearly once $n_\ell \cdot u \ge 1$, where the bounds become
useless (\cref{sec:deviation} shows where this bites).

\section{Bounding Deviation Between Real and Floating-Point Execution}\label{sec:deviation}

Throughout this section we assume that the conditions of \cref{thm:overflow-free} hold at each
layer, so that every floating-point operation is finite and the error model of
\cref{sec:float-error} applies.
We want a computable scalar $D_\ell$ for each layer $\ell = 1,\ldots,L-1$ such that
$\|d_\ell(x')\|_2 \le D_\ell$ for all $x' \in B(x,\varepsilon)$. The final layer
is treated separately in \cref{sec:robustness}, because what matters there is not
$\|d_L\|_2$ but the deviation in the output \emph{margins}, which needs a different analysis fed
by~$D_{L-1}$. The bounds $D_1,\ldots,D_{L-1}$ satisfy the two-coefficient linear recurrence
$D_\ell = \alpha_\ell\,D_{\ell-1} + \beta_\ell(r_{\ell-1})$, whose \emph{amplification factor}
$\alpha_\ell$ and \emph{fresh-error term} $\beta_\ell(r)$ depend only on the network weights and
the format constants; they come from a one-step bound (\cref{lem:one-step}) by substituting the
radii of \cref{sec:overflow}.

\paragraph{Setup}
Recall from \cref{eqn:mixed-model-matvec} that $\fl{W_\ell \hat{z}_{\ell-1}} = (W_\ell + \Delta W_\ell)\hat{z}_{\ell-1} + \etamv$
with $|\Delta W_\ell| \le \gamma_{n_\ell}|W_\ell|$ entrywise.  By the dominated-spectral-norm
inequality ($\|A\|_2 \le \|B\|_2$ whenever $|A_{ij}| \le B_{ij}$), this gives
$\|\Delta W_\ell v\|_2 \le \gamma_{n_\ell}\,\asnorm{W_\ell}\,\|v\|_2$ for any vector~$v$,
where $\asnorm{W_\ell}$ is the spectral norm of the
entrywise-absolute-value matrix~$|W_\ell|$; this is why that norm appears alongside
$\|W_\ell\|_2$ in everything that follows.

The following lemma bounds the one-step deviation $\|d_\ell\|_2$ in terms of the
previous-layer deviation $\|d_{\ell-1}\|_2$ and the activation norms $\|z_{\ell-1}\|_2$
and $\|\hat{z}_{\ell-1}\|_2$.  Its four-term form is unavoidable: each term corresponds
to a distinct source of rounding error in the layer computation. Three features of how those sources
appear are not visible from the error model alone. The weight-rounding perturbation~$\Delta W_\ell$ acts on $\hat{z}_{\ell-1} = z_{\ell-1} +
d_{\ell-1}$, so it splits across the two: that is why $\|d_{\ell-1}\|_2$ carries
$\gamma_{n_\ell}\asnorm{W_\ell}$ on top of $\|W_\ell\|_2$, and why $\|z_{\ell-1}\|_2$ appears with the
same factor. The bias-addition rounding is relative to the \emph{rounded} product, which is why the
third term is charged against the floating-point activation $\|\hat{z}_{\ell-1}\|_2$ rather than the
exact one, and why the absolute dot-product error picks up the factor $(1+u)$: the bias addition
rounds it once more.

\begin{lemma}[One-step deviation]\label[lemma]{lem:one-step}
Suppose the conditions of \cref{thm:overflow-free} hold at layer~$\ell$.
Then, for any $x' \in B(x,\varepsilon)$, with $d_\ell$, $d_{\ell-1}$, $z_{\ell-1}$,
and $\hat{z}_{\ell-1}$ denoting the respective quantities evaluated at input~$x'$,
\begin{equation}\label{eq:one-step}
\begin{aligned}
  \|d_\ell\|_2 \;\le\;& L(\varphi_\ell)\Bigl[
    \bigl(\|W_\ell\|_2 + \gamma_{n_\ell}\,\asnorm{W_\ell}\bigr)\,\|d_{\ell-1}\|_2 \\
  &\qquad +\; \gamma_{n_\ell}\,\asnorm{W_\ell}\,\|z_{\ell-1}\|_2 \\
  &\qquad +\; u\bigl((1+\gamma_{n_\ell})\,\asnorm{W_\ell}\,\|\hat{z}_{\ell-1}\|_2
                      + \|b_\ell\|_2\bigr) \\
  &\qquad +\; (1+u)\,\adot{n_\ell}\,\sqrt{m_\ell}\Bigr].
\end{aligned}
\end{equation}
\end{lemma}

\paragraph{Linear recursion}
The lemma involves the exact and the floating-point activation norms. Substituting the radius
bound $\|z_{\ell-1}\|_2 \le r_{\ell-1}$ of \cref{sec:overflow} for the first and, by the
triangle inequality, $r_{\ell-1} + \|d_{\ell-1}\|_2$ for the second, then grouping by
coefficient of $\|d_{\ell-1}\|_2$ versus $r_{\ell-1}$, collapses the
four terms into two. The grouping produces
\[
  \kappa_{n_\ell} \;\defeq\; \gamma_{n_\ell} + u(1+\gamma_{n_\ell}),
\]
which combines the relative rounding factor $\gamma_{n_\ell}$ of the matrix-vector product
with the factor $u(1+\gamma_{n_\ell})$ from the bias addition's rounding of that product, and
\begin{align*}
  \alpha_\ell &\;\defeq\; L(\varphi_\ell)\,\bigl(\|W_\ell\|_2 + \kappa_{n_\ell}\,\asnorm{W_\ell}\bigr), \\
  \beta_\ell(r) &\;\defeq\; L(\varphi_\ell)\,\Bigl(\kappa_{n_\ell}\,\asnorm{W_\ell}\cdot r
                   \;+\; u\,\|b_\ell\|_2
                   \;+\; (1+u)\,\adot{n_\ell}\,\sqrt{m_\ell}\Bigr).
\end{align*}

\begin{corollary}[Linear recursion for deviation]\label[corollary]{cor:recursion}
For every $x' \in B(x,\varepsilon)$ and every layer $\ell \ge 1$,
\[
  \|d_\ell(x')\|_2 \;\le\; \alpha_\ell\,\|d_{\ell-1}(x')\|_2 \;+\; \beta_\ell(r_{\ell-1}).
\]
\end{corollary}

Like the radii of \cref{sec:overflow}, the coefficients mirror the layer computation, with the
Lipschitz factor now inside them: the amplification factor is the most a layer can scale an
incoming deviation, $\|W_\ell\|_2$ with the floating-point correction
$\kappa_{n_\ell}\,\asnorm{W_\ell}$, and the fresh-error term is the layer's own rounding error,
dominated by $\kappa_{n_\ell}\,\asnorm{W_\ell}\,r_{\ell-1}$ because that error
grows with the activations it rounds. At float32 and float64, for the ReLU networks
($L(\varphi_\ell) = 1$) of the scale we evaluate, both are only marginally perturbed from what
they approximate: $\alpha_\ell$ is essentially $\|W_\ell\|_2$ and
$\beta_\ell(r_{\ell-1}) \ll r_{\ell-1}$. At float16, however, $\kappa_{n_\ell}$ inflates even for
these networks, both coefficients grow, and the accumulated deviation $D_{L-1}$ grows past the
point at which certification is feasible (\cref{sec:eval}).

\paragraph{Cumulative deviation bound}
With $D_0 \defeq 0$, since the input is used exactly (\cref{sec:overflow}), define
\[
  D_\ell \;\defeq\; \alpha_\ell\,D_{\ell-1} \;+\; \beta_\ell(r_{\ell-1}), \qquad \ell = 1,\ldots,L-1;
\]
by induction on \cref{cor:recursion}, $\|d_\ell(x')\|_2 \le D_\ell$ simultaneously for all
$x' \in B(x,\varepsilon)$. These are the bounds that the overflow checks of \cref{sec:overflow}
consume.

The radii enter the recursion only through $\beta_\ell(r_{\ell-1})$, so any sequence
satisfying $\|z_\ell(x')\|_2 \le r_\ell$ for all $x' \in B(x,\varepsilon)$ may replace the
deterministic radii of \cref{sec:overflow}; tighter, input-specific radii reduce the deviation
bounds, and \cref{sec:hybrid} obtains such radii by measuring a high-precision execution.

\section{Robustness Certification under Floating-Point Execution}\label{sec:robustness}

An input~$x$ is robust under floating-point execution when every floating-point margin
$\hat{m}_{j,i^*}(x') = \hat{y}_{i^*}(x') - \hat{y}_j(x')$ remains positive across
$B(x,\varepsilon)$ for every competing class~$j \ne i^*$, where $i^*$ is the class predicted
at~$x$. This margin is the exact difference of two floating-point outputs (the outputs are compared
by $\ArgMax$, not subtracted in floating point), each output $\hat{y}_i$ the separately
rounded row $\fl{\fl{W_{L,i} \cdot \hat{z}_{L-1}} + b_{L,i}}$, the final layer having identity
activation. To certify this we bound how far $\hat{m}_{j,i^*}$ departs from the
real-arithmetic margin $m_{j,i^*}$ of \cref{sec:lipschitz}, the quantity that the classical
check~\eqref{eqn:certification-check} keeps positive across the ball, and which, as the same two
rows computed exactly, is
\[
  m_{j,i^*} \;=\; (W_{L,i^*} - W_{L,j}) \cdot z_{L-1} + (b_{L,i^*} - b_{L,j}),
\]
where $W_{L,i^*}$ and $W_{L,j}$ denote rows $i^*$ and~$j$ of~$W_L$.
\label{sec:D}

\begin{lemma}[Final-layer pairwise deviation]\label[lemma]{lem:final-deviation}
Let $\hat{z}_{L-1}$ satisfy $\|\hat{z}_{L-1} - z_{L-1}\|_2 \le D_{L-1}$ and
$\|z_{L-1}\|_2 \le r_{L-1}$. Suppose the conditions of \cref{thm:overflow-free} hold at
layer~$L$. Then
\[
  |\hat{m}_{j,i^*} - m_{j,i^*}| \;\le\;
  \alpha_L^{(j,i^*)} \cdot D_{L-1} + \beta_L^{(j,i^*)}(r_{L-1}),
\]
where
\begin{align*}
  \alpha_L^{(j,i^*)} &\;\defeq\; \|W_{L,i^*} - W_{L,j}\|_2
                      + \kappa_{n_L} \cdot \bigl\||W_{L,i^*}| + |W_{L,j}|\bigr\|_2, \\
  \beta_L^{(j,i^*)}(r) &\;\defeq\; \kappa_{n_L} \cdot \bigl\||W_{L,i^*}| + |W_{L,j}|\bigr\|_2 \cdot r \\
                       &\qquad +\; u \cdot (|b_{L,i^*}| + |b_{L,j}|)
                       + 2(1+u) \cdot \adot{n_L}.
\end{align*}
\end{lemma}
The shape is that of \cref{lem:one-step} applied to the difference of two rows, with one
departure: the rounding perturbation is bounded by
$\kappa_{n_L}(|W_{L,i^*}| + |W_{L,j}|)$, not the smaller $\kappa_{n_L}|W_{L,i^*} - W_{L,j}|$, because
the two rows' dot products are rounded independently and their errors need not cancel.

\paragraph{Two executions, two error bounds}
The certificate compares two executions: at the centre~$x$, where the floating-point margin
$\hat{m}_{j,i^*}(x)$ is observed, and at an arbitrary $x' \in B(x,\varepsilon)$, where
$\hat{m}_{j,i^*}(x') > 0$ must be guaranteed. We write
$r_\ell(x,\varepsilon)$ and $D_\ell(x,\varepsilon)$ to make the input and the radius explicit
(they were implicit in \cref{sec:overflow,sec:deviation}). The centre is the ball of radius~$0$,
so its recurrences start from $r_0 = \|x\|_2$ and give $D_{\text{ctr}} \defeq D_{L-1}(x,0)$ and
$r_{\text{ctr}} \defeq r_{L-1}(x,0)$; the ball starts from $r_0 = \|x\|_2 + \varepsilon$ and
gives $D_{\text{ball}} \defeq D_{L-1}(x,\varepsilon)$ and
$r_{\text{ball}} \defeq r_{L-1}(x,\varepsilon)$. The two error bounds are
\[
  E_{\text{ctr}}^{(j,i^*)} \;\defeq\;
  \alpha_L^{(j,i^*)} \cdot D_{\text{ctr}} + \beta_L^{(j,i^*)}(r_{\text{ctr}}),
  \qquad
  E_{\text{ball}}^{(j,i^*)} \;\defeq\;
  \alpha_L^{(j,i^*)} \cdot D_{\text{ball}} + \beta_L^{(j,i^*)}(r_{\text{ball}}).
\]

Why both are needed is the chain of three inequalities behind the certificate. For
$x' \in B(x,\varepsilon)$ and a Lipschitz constant $L_{j,i^*}$ of the real-arithmetic margin on
the ball,
\begin{align*}
  \hat{m}_{j,i^*}(x')
  &\;\ge\; m_{j,i^*}(x') - E_{\text{ball}}^{(j,i^*)}
  && \text{(\cref{lem:final-deviation} at~$x'$)} \\
  &\;\ge\; m_{j,i^*}(x) - L_{j,i^*}\,\varepsilon - E_{\text{ball}}^{(j,i^*)}
  && \text{(Lipschitz, in real arithmetic)} \\
  &\;\ge\; \hat{m}_{j,i^*}(x) - E_{\text{ctr}}^{(j,i^*)} - L_{j,i^*}\,\varepsilon - E_{\text{ball}}^{(j,i^*)}
  && \text{(\cref{lem:final-deviation} at~$x$)}.
\end{align*}
The Lipschitz step is available only for the real-arithmetic margin, since the floating-point
execution is not a Lipschitz function (\cref{sec:intro}), so the argument must cross from floating
point to real arithmetic at~$x'$ and back at~$x$, paying a margin deviation each time. Requiring
the last line to be positive is the certification condition.

\begin{theorem}[Floating-point robustness certificate]\label{thm:fp-robust}
Let $x$ be a format-representable input, $\varepsilon \ge 0$, and $i^*$ a class. Let the
network satisfy the assumptions of \cref{sec:float-semantics} (in particular FP-exact activations,
assumption~\ref{assumption-fp-exact}, and weights and biases finite in the deployment format),
with identity activation at the final layer. Suppose:
\begin{enumerate}[label=(\roman*)]
  \item The conditions of \cref{thm:overflow-free} hold at every layer with
    $r_0 = \|x\|_2 + \varepsilon$.
  \item For each competing class $j \ne i^*$, there exists a margin Lipschitz constant
    $L_{j,i^*} \ge 0$ valid on the ball: for all $x_1, x_2 \in B(x,\varepsilon)$,
    \[
      |m_{j,i^*}(x_1) - m_{j,i^*}(x_2)| \;\le\; L_{j,i^*} \cdot \|x_1 - x_2\|_2.
    \]
  \item The \emph{certification condition} holds for each $j \ne i^*$, where
    $E_{\text{ctr}}^{(j,i^*)}$ and $E_{\text{ball}}^{(j,i^*)}$ are the error bounds defined above:
    \[
      \hat{m}_{j,i^*}(x) - L_{j,i^*} \cdot \varepsilon - (E_{\text{ctr}}^{(j,i^*)} + E_{\text{ball}}^{(j,i^*)})
      \;>\; 0.
    \]
\end{enumerate}
Then for every format-representable $x' \in B(x,\varepsilon)$, the floating-point network
executes without overflow and its margin satisfies $\hat{m}_{j,i^*}(x') > 0$ for every
$j \ne i^*$; hence $\ArgMax(\hat{N}(x')) = i^*$.
\end{theorem}
Condition~(ii) asks only for a Lipschitz constant valid on $B(x,\varepsilon)$: the global
constant $L_{j,i^*}$ of \cref{sec:lipschitz}, which we use, is one admissible choice and a local
constant valid on the ball is another; conditions~(i) and~(iii) are unaffected by the choice.

Passing the check implies the classical check~\eqref{eqn:certification-check}, since
\cref{lem:final-deviation} at the centre gives
$m_{j,i^*}(x) \ge \hat{m}_{j,i^*}(x) - E_{\text{ctr}}^{(j,i^*)}$ and
$E_{\text{ball}}^{(j,i^*)} \ge 0$. The added term
$E_{\text{ctr}}^{(j,i^*)} + E_{\text{ball}}^{(j,i^*)}$ is thus the \emph{margin degradation}
due to floating-point error, which \cref{sec:eval} quantifies.

\section{Pre-Deployment Hybrid Certification}\label{sec:hybrid}

The deviation bounds $D_\ell$ of \cref{sec:deviation} are worst-case: they hold for every
$x' \in B(x,\varepsilon)$ and are computed from the network weights and the deterministic radii
alone. At the deployment format they compound worst cases through the layers and can far exceed
the deviation the network actually incurs (\cref{sec:eval}); at a high-precision format such as
float64 the same bounds are negligible in absolute terms. This section targets the
\emph{pre-deployment} setting, in which robustness verifiers are run and their results reported.
Before a model is deployed, one is free to execute it in any format, at any input, and to take
whatever measurements of its behaviour might aid certification. We show how measurements of a float64 execution, combined with
that execution's own tight deviation bounds, bound the deployment format's deviation far more
tightly than the worst-case~$D_\ell$.

At the penultimate layer~$L-1$, one can measure how far the deployment-format activation at an
input~$x$ lies from that of a high-precision execution, float64 in our case. For this measurement
to bound the deployment format's deviation $d_{L-1}(x)$ from real arithmetic soundly, the
high-precision network must represent the deployment format's weights and biases exactly
(float64 represents every float32 value) and must be checked to have finite preactivations, so
that the bound of \cref{sec:deviation} applies to it. Writing the superscript $\mathrm{hi}$ for
the high-precision execution, the \emph{hybrid} deviation at~$x$ is the measurement plus that
execution's own worst-case deviation bound $D^{\mathrm{hi}}_{L-1}(x,0)$:
\[
D_{L-1}^\mathrm{hybrid}(x) \defeq \| \hat{z}_{L-1}(x) - \hat{z}_{L-1}^\mathrm{hi}(x)\|_2 + D_{L-1}^\mathrm{hi}(x,0) .
\]
Since $\|d_{L-1}^\mathrm{hi}\|_2 \leq D_{L-1}^\mathrm{hi}$ by \cref{cor:recursion}, the triangle
inequality gives:
\begin{lemma}[Hybrid deviation bound]\label[lemma]{lem:hybrid}
  For an input point~$x$:
  \[
  \|d_{L-1}(x)\|_2 \le D_{L-1}^\mathrm{hybrid}(x)
  \]
\end{lemma}

\Cref{lem:final-deviation} at the single point~$x$ (the ball of radius~$0$) therefore holds with
$D_{L-1}^\mathrm{hybrid}(x)$ in place of $D_{L-1}$, so in the certification check of
\cref{thm:fp-robust} the centre error $E_\mathrm{ctr}^{(j,i^*)}$ may be replaced by
\[
      E_{\text{ctr}}^{\mathrm{hybrid},(j,i^*)} \;\defeq\;
      \alpha_L^{(j,i^*)} \cdot D_{L-1}^\mathrm{hybrid}(x) + \beta_L^{(j,i^*)}(r_{\text{ctr}}) .
\]

\begin{theorem}[Hybrid-Centre certificate]\label{thm:fp-robust-hybrid}
  Let the high-precision network be the deployed network with its weights, biases and
  activations unchanged, executed in a higher-precision format in which the deployment
  format's values are representable, and let its forward pass at~$x$ produce finite
  preactivations at every layer.
  Let the conditions of \cref{thm:fp-robust} hold for the input point~$x$, except the certification check is replaced by:
    \[
      \hat{m}_{j,i^*}(x) - L_{j,i^*} \cdot \varepsilon - (E_{\text{ctr}}^{\mathrm{hybrid},(j,i^*)} + E_{\text{ball}}^{(j,i^*)})
      \;>\; 0.
      \]
    Then, as in \cref{thm:fp-robust}, $x$ is robust for $\varepsilon$.
\end{theorem}

The same high-precision execution can also tighten the ball error $E_\mathrm{ball}^{(j,i^*)}$. That
term covers the whole ball $B(x,\varepsilon)$, so it cannot be measured at the single point~$x$.
Its conservatism sits in the deterministic radii $r_\ell$ of \cref{sec:radii}, which propagate the
input radius $r_0 = \|x\|_2 + \varepsilon$ through the layerwise Lipschitz constants, a worst case
that typically grows the radii far beyond the activation norms the network realises. Since the
deviation recursion of \cref{cor:recursion} accepts any radii that bound the activation norms over
the ball (\cref{sec:deviation}), we compute tighter ones from the high-precision execution. At the centre,
the high-precision activation norm $\|\hat{z}_\ell^\mathrm{hi}(x)\|_2$ is a measured proxy for the
exact norm $\|z_\ell(x)\|_2$, accurate up to $D_\ell^\mathrm{hi}(x,0)$. To cover the ball we inflate
it by the most any $x' \in B(x,\varepsilon)$ can move the layer-$\ell$ activation: $\varepsilon$
times the Lipschitz constant $\prod_{k=1}^{\ell} L(\varphi_k)\|W_k\|_2$ of the map from the input
to layer~$\ell$. The \emph{measured radius} is thus the measurement, the ball's reach, and the
measurement's own error:
\[
  r_\ell^\mathrm{meas}(x,\varepsilon) \;\defeq\;
  \|\hat{z}_\ell^\mathrm{hi}(x)\|_2
  \;+\; \Big(\textstyle\prod_{k=1}^{\ell} L(\varphi_k)\|W_k\|_2\Big)\,\varepsilon
  \;+\; D_\ell^\mathrm{hi}(x,0) .
\]

\begin{lemma}[Measured radii]\label[lemma]{lem:measured-radii}
  For all $x' \in B(x,\varepsilon)$ and every layer~$\ell$, $\|z_\ell(x')\|_2 \le r_\ell^\mathrm{meas}(x,\varepsilon)$.
\end{lemma}

Running the deviation recursion of \cref{cor:recursion} with $r_\ell^\mathrm{meas}(x,\varepsilon)$
in place of $r_\ell$ therefore yields bounds $D_{\ell}^\mathrm{meas}(x,\varepsilon)$ on
$\|d_{\ell}(x')\|_2$ at every layer, valid for all $x' \in B(x,\varepsilon)$. For the ball error this replaces both
the deviation $D_{L-1}$ and the radius in the fresh-error term by their measured counterparts,
\[
  E_{\text{ball}}^{\mathrm{meas},(j,i^*)} \;\defeq\;
  \alpha_L^{(j,i^*)} \cdot D_{L-1}^\mathrm{meas}(x,\varepsilon)
  \;+\; \beta_L^{(j,i^*)}\!\big(r_{L-1}^\mathrm{meas}(x,\varepsilon)\big).
\]
For the centre error the deviation is already measured, via $D_{L-1}^\mathrm{hybrid}(x)$ of
\cref{lem:hybrid}; the measured radii additionally sharpen its fresh-error term, giving
\[
  E_{\text{ctr}}^{\mathrm{meas},(j,i^*)} \;\defeq\;
  \alpha_L^{(j,i^*)} \cdot D_{L-1}^\mathrm{hybrid}(x)
  \;+\; \beta_L^{(j,i^*)}\!\big(r_{L-1}^\mathrm{meas}(x,0)\big).
\]

\begin{theorem}[Measured-Radii certificate]\label{thm:fp-robust-hybrid-meas}
  Let the conditions of \cref{thm:fp-robust-hybrid} hold, including the overflow conditions~(i)
  of \cref{thm:fp-robust} at the deterministic radii, except that $E_{\text{ctr}}^{\mathrm{hybrid},(j,i^*)}$
  and $E_{\text{ball}}^{(j,i^*)}$ are replaced by $E_{\text{ctr}}^{\mathrm{meas},(j,i^*)}$ and
  $E_{\text{ball}}^{\mathrm{meas},(j,i^*)}$ in the certification check. Then, as before, $x$ is
  robust for~$\varepsilon$.
\end{theorem}
Our certifier checks the overflow conditions~(i) at the deterministic radii, which need no reference
execution; they pass comfortably for every model we evaluate at float32 and float64
(\cref{sec:eval-setup}). A single high-precision pass supplies both the centre measurement and
the per-layer norms; how much the measured radii gain over the deterministic ones depends on how
far the realised activation norms fall short of their Lipschitz-product bound, which
\cref{sec:eval} quantifies.

\section{Sound and Scalable Spectral-Norm Computation}\label{sec:norms}

The certificate of \cref{sec:robustness} and the deviation and overflow bounds of
\cref{sec:deviation,sec:overflow} all require sound \emph{upper} bounds on the per-layer
spectral norms $\|W_\ell\|_2$ and $\asnorm{W_\ell}$; both reduce to the single problem of
bounding $\|M\|_2$ from above for a matrix~$M$ of shape $m_\ell \times n_\ell$ (\cref{sec:nn-model}).
Tobler et al.~\cite{tobler2025} keep rounding out of the certifier's own computation, the second
place where it bears on certification (\cref{sec:intro}), by working in exact rational arithmetic
throughout, at significant cost: the spectral norms are
computed once per model, before any input is certified, but even as a one-off precomputation they
take tens of hours for CIFAR-10's input layer ($n_1 = 3072$). We remove that cost by
running the dominant computation in float64 and bounding its rounding with the error model
the preceding sections apply to the network, so we still obtain a sound rational upper bound
(mechanised in Rocq). Our implementation keeps exact rational arithmetic everywhere else, and its
precomputation drops from tens of hours to minutes (\cref{sec:rq2}).

A sound upper bound on $\|M\|_2$ comes for free from the classical closed
forms~\cite[Ch.~6]{higham2002}: the Frobenius norm $\|M\|_F$, the square root of the sum of $M$'s
squared entries, and $\sqrt{\|M\|_1\|M\|_\infty}$. But they are loose: on our models they
over-estimate each hidden layer's norm by $4$--$16\times$, inflating the product over layers by
$10^{4}$--$10^{10}$, which would make every certificate vacuous. \emph{Gram
iteration}~\cite{delattre2023}, which Tobler et al.~\cite{tobler2025} use, tightens the Frobenius
bound while keeping it an upper bound. The algorithm computes the \emph{iterates} $A_0 = M$
and $A_k = A_{k-1}^\top A_{k-1}$, each the Gram matrix of the one before.
Since $\|X^\top X\|_2 = \|X\|_2^2$, each iterate's spectral norm is the \emph{square} of
its predecessor's, so $\|M\|_2 = \|A_n\|_2^{1/2^n}$; since $\|X\|_2 \le \|X\|_F$, bounding
$\|A_n\|_2$ by $\|A_n\|_F$ gives $\|M\|_2 \le \|A_n\|_F^{1/2^n}$, recovered by $n$ square roots. The
bound $\|A_n\|_F^{1/2^n}$ tightens superlinearly in~$n$~\cite{delattre2023}. To keep the entries from overflowing, each iterate is rescaled by a bound on its
Frobenius norm before the next product, and the rescalings are undone alongside the square roots.

Each rescaled iterate is also \emph{truncated} before the next product: in Tobler et al.'s
procedure~\cite{tobler2025} to bounded precision, in ours, after the rational rescaling, back to
float64, the format the product runs in. The truncation error is measured and, by the triangle inequality for the spectral norm, added
to the running bound, so the result stays an upper bound.

Both procedures rest on three sound rational over-approximations. The rounded-up square root
$\SqrtUB(x)\ge\sqrt{x}$ (Heron's method) we inherit unchanged. The other two are ours, float64
counterparts of Tobler et al.'s rational ones. The truncation $\Truncate(B)$ rounds~$B$ to the
nearest float64 matrix~$A$ and returns a rational bound $t\ge\|B-A\|_F$ on the rounding. The
Frobenius-norm upper bound $\FrobUB(X)\ge\|X\|_F$ sums the squares of $X$'s entries in float64,
inflates the sum by the dot-product bound behind \cref{lem:forward-error}, and takes a rounded-up
square root. That sum is a dot product with one term per entry of~$X$, millions of terms for our
matrices, and at that length the exact $\gamma_n$ is prohibitively expensive to materialise, so
the inflation uses the closed form $\bar\gamma_n \defeq nu/(1-nu) \ge \gamma_n$ in its place
(\cref{app:derivations} defines $\FrobUB$ in full).

\paragraph{A float64 adaptation}
The dominant cost is the Gram product, cubic in~$n_\ell$; running it in hardware float64 replaces each
exact-rational product by a float64 sum of products at constant cost per operation, independent
of the entries' magnitude. The product is now itself inexact, so we bound its Frobenius error after checking that every entry of the product is finite:
\begin{lemma}[Frobenius error of a floating-point Gram product]\label[lemma]{lem:gram-fp}
Let $A$ be a matrix of floating-point entries with $p$ rows and $q$ columns, and let
$\tilde N \defeq \fl{A^\top A}$ be the floating-point evaluation of $A^\top A$, formed from dot products of
length~$p$. If every entry of $\tilde N$ is finite, then
\[
  \|\tilde N - A^\top A\|_F \;\le\; \xi(A),\quad \textrm{\emph{where we define\ }}\ \xi(A) \;\defeq\;
  \gamma_p\,\FrobUB(A)^2 \;+\; \adotfwd{p}\cdot q.
\]
\end{lemma}

The terms $\gamma_p$ and $\adotfwd{p}$ are the constants of \cref{lem:forward-error}, so
$\xi(A)$ is computed from $\FrobUB(A)$ alone. Its two terms are the relative and absolute parts
of the dot-product error bound behind that lemma over the $q^2$ entries of~$\tilde N$, each a length-$p$ dot product of two columns
of~$A$: the relative parts, each at most $\gamma_p$ times the product of its two columns' norms,
combine under the Frobenius norm into $\gamma_p\|A\|_F^2$, and the absolute parts, $\adotfwd{p}$
each, into $\adotfwd{p}\,q$.

\begin{algorithm}
\caption{$\mathrm{GramIterFP}(M,n)$: returns a rational $s \ge \|M\|_2$. Same control structure
as Tobler et al.'s~\cite{tobler2025} error-tracked Gram iteration; the float64 adaptation is
described in the text. $\FrobUB$, $\SqrtUB$ and $\Truncate$ are as defined in the text.}
\label{alg:gramfp}
\begin{algorithmic}[1]
\State $A \gets M$ \Comment{float64 (float32 weights are exactly representable)}
\For{$k \gets 1, \ldots, n$}
  \State $\tilde N \gets \fl{A^\top A}$ \Comment{float64 Gram product; must be finite}
  \State $c_k \gets \FrobUB(\tilde N)$;\enspace $\xi_k \gets \xi(A)$ \Comment{rescale factor; product error (\cref{lem:gram-fp})}
  \State $(A, t_k) \gets \Truncate(\tilde N / c_k)$ \Comment{rescale, truncate to float64}
  \State $\delta_k \gets t_k + \xi_k / c_k$ \Comment{per-step error}
\EndFor
\State $s \gets \FrobUB(A)$
\For{$k \gets n, \ldots, 1$}
  \State $s \gets \SqrtUB\bigl(c_k\,(s + \delta_k)\bigr)$ \Comment{undo rescaling and squaring}
\EndFor
\State \Return $s$
\end{algorithmic}
\end{algorithm}

In \cref{alg:gramfp}, only the $O(n_\ell^2)$ error bookkeeping stays in exact rational
arithmetic (on bounded-bit-length numbers), while the $O(n_\ell^3)$ product is float64. Each step's error
$\delta_k = t_k + \xi_k/c_k$ combines the truncation error $t_k$ with the product error of
\cref{lem:gram-fp}, the latter divided by the rescaling factor $c_k$ because it is incurred
before the rescaling (line~5), unlike the truncation error; the backward pass propagates the
accumulated $\delta_k$ to return a rational $s \ge \|M\|_2$.

Not just \cref{lem:gram-fp} but the full procedure is mechanised, for every run whose runtime
checks pass; a failing check refuses to certify (\cref{app:derivations} states the checks).

Since $\|M\|_2 = \|M^\top\|_2$, the iteration may moreover run on whichever of
$M^\top M$ ($n_\ell \times n_\ell$) and $M M^\top$ ($m_\ell \times m_\ell$) is smaller, sound by
the same argument applied to $M^\top$. For CIFAR-10's input layer ($m_1 = 512$, $n_1 = 3072$)
this brings minutes down to seconds.

\section{Evaluation}\label{sec:eval}

We implement a certifier that embodies the certification checks of
\cref{thm:fp-robust,thm:fp-robust-hybrid,thm:fp-robust-hybrid-meas} (\cref{sec:eval-setup}) and
evaluate its performance empirically around three research questions.
\textbf{RQ1} validates it on the known counterexamples of \cref{sec:cex}: does it reject every
false certificate identified there?
\textbf{RQ2} measures the cost of floating-point soundness, the drop in certified robustness
relative to real arithmetic, across image, tabular and many-class tasks, and asks what governs
that cost where it is large, network scale or the certification task. It then asks how much of
that cost the pre-deployment modes of \cref{sec:hybrid} reclaim, and reports the time to
certify.
\textbf{RQ3} asks whether floating-point soundness at float32 costs the two tools alike,
comparing the certificates of every mode and their certification time with ERAN's~\cite{ERAN}.

\begin{table}
\centering
\small
\setlength{\tabcolsep}{3pt}
\begin{tabular}{l c r c cccc cccc}
\toprule
 & & & & \multicolumn{4}{c}{Robustness (\%)} & \multicolumn{4}{c}{VRA (\%)} \\
\cmidrule(lr){5-8}\cmidrule(lr){9-12}
Model & $\varepsilon$ & $N$ & Acc & Real & Std & Hyb & Meas & Real & Std & Hyb & Meas \\
\midrule
MNIST           & 0.3   & 10{,}000   & 98.40 & 95.74 & 95.50 & 95.62 & 95.65 & 95.40 & 95.17 & 95.28 & 95.31 \\
Fashion MNIST   & 0.25  & 10{,}000   & 89.10 & 83.65 & 82.16 & 82.80 & 82.98 & 79.54 & 78.47 & 78.94 & 79.08 \\
CIFAR-10        & 0.141 & 10{,}000   & 57.74 & 46.12 & 18.90 & 30.14 & 32.95 & 35.95 & 16.96 & 25.57 & 27.52 \\
HIGGS-1024      & 0.1   & 500{,}000  & 71.91 & 82.88 & 76.76 & 78.21 & 81.62 & 62.69 & 59.13 & 59.98 & 61.97 \\
EMNIST-ByClass  & 0.3   & 116{,}323  & 83.07 & 86.17 & 85.01 & 85.59 & 85.66 & 75.37 & 74.54 & 74.97 & 75.01 \\
EMNIST-Balanced & 0.3   & 18{,}800   & 83.59 & 77.75 & 74.06 & 75.80 & 76.24 & 71.25 & 68.33 & 69.74 & 70.06 \\
\bottomrule
\end{tabular}
\caption{\label{tab:master}%
  Certification on each model's full test set (float32; 12 Gram iterations). \emph{Real} is the
  real-arithmetic ceiling; \emph{Std}, \emph{Hyb} and \emph{Meas} are the Standard, Hybrid-Centre
  and Measured-Radii modes of \cref{sec:eval-setup}. \emph{Robustness} is the certified rate and
  \emph{VRA} the verified robust accuracy (\cref{sec:eval-setup}); \emph{Acc} is clean test
  accuracy, the ceiling on VRA; $N$ is the test-set size. For HIGGS-1024, $\varepsilon$ is in
  standardised units (\cref{app:training}).}
\end{table}

\subsection{Experimental Setup}\label{sec:eval-setup}

Our certifier ran on an Apple M4 laptop (10-core CPU, 16\,GB RAM) under macOS~26.6.1. Models
were executed by a single-threaded, per-operation numpy CPU forward pass, not through TensorFlow, whose float32 path flushes subnormals to zero (\cref{app:ftz}). We validated empirically at each format that this execution conforms to
the semantics our theory assumes (\cref{sec:float-error}).

\paragraph{Implementation and trust boundary}
Our certifier extends a re-implementation of Tobler et al.'s real-arithmetic
certifier~\cite{tobler2025} with the floating-point-sound components of
\cref{sec:overflow,sec:deviation,sec:robustness,sec:hybrid,sec:norms}, in about 2,000 lines of
Python. Like theirs, it computes every bound in exact rational arithmetic (the gmpy2 library),
the one exception being the spectral-norm computation, implemented as \cref{sec:norms} describes.
The theory is mechanised in Rocq, including every numbered result. The certifier's
implementation is not mechanised in Rocq. It is instead validated: its
real-arithmetic kernel exactly reproduces the reference outputs of Tobler et al.'s verified
certifier~\cite{tobler2025} on their three image models, and the floating-point extensions are
carefully audited. Verified extraction of the certifier would close this final gap; it remains
future work. The certifier refuses any model with a layer at which $n_\ell u \ge 1$, its bounds
there being sound but useless (\cref{sec:nell-cdot-u-lt-1}); this excludes CIFAR-10 at float16.

The certifier operates in three modes, abbreviated \emph{Std}, \emph{Hyb} and \emph{Meas} in the
tables. \textbf{Standard} (\cref{thm:fp-robust}) certifies from the worst-case radii and deviation
bounds alone. The two \emph{pre-deployment} modes also consult a float64 execution of the model: \textbf{Hybrid-Centre} (\cref{thm:fp-robust-hybrid}) uses the measured centre deviation,
and \textbf{Measured-Radii} (\cref{thm:fp-robust-hybrid-meas}) additionally the measured radii.

\paragraph{Models and datasets}
We evaluate on six globally robust dense ReLU classifiers. Three are the image classifiers of
Tobler et al.~\cite{tobler2025}: MNIST~\cite{MNIST}, Fashion MNIST~\cite{Fashion_MNIST} and
CIFAR-10~\cite{CIFAR10}, with 784- or 3072-dimensional inputs, ten classes, and eight to twelve
hidden layers of 128 to 512 neurons. To probe two task properties those benchmarks do not vary,
we train, with globally robust training~\cite{leino2021}, classifiers for a \emph{tabular} task
and for \emph{many-class} tasks. HIGGS~\cite{baldi2014} is a binary high-energy-physics
benchmark with 28 standardised continuous features; we train it as a \emph{width sweep},
five-hidden-layer networks at widths 128, 256, 512 and 1024, so that \textbf{RQ2} can isolate the
effect of network width at a fixed task. EMNIST~\cite{cohen2017emnist} is handwritten-character
recognition, an image task like MNIST but with many classes; we train its two standard splits,
ByClass (62 classes, imbalanced) and Balanced (47 classes, merging visually similar letter
cases, with equal class sizes).
EMNIST-ByClass deliberately reuses CIFAR-10's architecture, giving a task-versus-architecture
control (\textbf{RQ2}). All networks have ReLU hidden activations and identity output and execute
in float32; \cref{app:training} gives every architecture and the training configuration of the
models we trained. Per-model radii~$\varepsilon$ and test-set sizes~$N$ are in \cref{tab:master}:
each is the model's full standard test set, except that the width sweep subsamples
$10{,}000$ HIGGS instances per width.

\paragraph{Spectral norms}\label{sec:gram-counts}
The bounds on $\|W_\ell\|_2$ and $\big\||W_\ell|\big\|_2$ come from \cref{alg:gramfp} with 12
iterations, one past convergence (relative tolerance $10^{-3}$) for every model;
\textbf{RQ1} uses tighter per-model counts, to minimise norm-induced conservatism on the
counterexamples (\cref{app:training}).

\paragraph{Floating-point formats}
Float32 is the deployment format throughout unless stated otherwise, and every model was trained in it.
At float64 rounding costs neither our tool nor ERAN anything. At float16
certification is vacuous in every mode, certifying no instance, as the condition
$n_\ell u \ll 1$ of \cref{sec:nell-cdot-u-lt-1} anticipates; so is ERAN (see below).

\paragraph{Metrics}
We report two quantities (\cref{tab:master}). \emph{Robustness} is the certified rate, the
fraction of test points certified robust, which measures the certifier's conservatism
independently of model accuracy. \emph{Verified robust accuracy} (VRA), the fraction of points
both certified robust and correctly classified, is the quantity of interest when certifying a
model pre-deployment. The
\emph{floating-point cost} is the drop from real-arithmetic to floating-point-sound
certification on either metric.

\paragraph{Baseline verifier}
\textbf{RQ3} compares against ERAN~\cite{ERAN}, to our knowledge the only publicly available
floating-point-sound verifier demonstrated on networks of our size, using its GPUPoly
domain~\cite{GPUPoly}, the GPU implementation of DeepPoly~\cite{DeepPoly}, run on one NVIDIA
A10 GPU (a Lambda Labs cloud instance; Ubuntu 22.04, CUDA 12.8) over the full test set of every
image model; HIGGS is excluded because ERAN's input pipeline is image-specific. ERAN
targets $\ell_\infty$ robustness rather than our $\ell_2$ robustness; we convert our radius via
$\varepsilon_{\ell_\infty} = \varepsilon_{\ell_2}/\sqrt{n_\mathrm{in}}$, rounded down, where
$n_\mathrm{in}$ is the input dimension. The resulting $\ell_\infty$ box lies within our
$\ell_2$ ball, so ERAN certifies a smaller perturbation set, a weaker guarantee
(an $\ell_\infty$ version of our certificate is future work, \cref{sec:conclusion}).

As shipped, GPUPoly certifies a float64 execution with a double-precision analysis (its
\emph{level~3}) or a float32 execution with a single-precision analysis (\emph{level~4}).
Level~4 is near-vacuous on four of our five models (the \emph{level~4} column
of \cref{tab:eran}). We
compare instead against its double-precision analysis with the rounding of its products charged
at the round-to-nearest bound of a chosen target format rather than at the double precision of
its analysis (\cref{app:gpupoly}). We call this configuration
the \emph{comparator at} that format; the \emph{comparator} alone is the comparator at float32,
our deployment format. With the products charged nothing at all, so that the only rounding left is the
analysis's own double-precision arithmetic, soundly directed outward, the analysis is the abstraction's real-arithmetic ceiling
(\emph{Real}), whose results coincide with level~3 and the comparator at float64 (\cref{app:gpupoly}).
The comparator (at float32) covers additions only to double precision and does
not charge the rounding of the input to the target format, so at float32 it under-charges the
rounding and is not a sound float32 verifier: it is intentionally generous to GPUPoly.
The comparator at float16 verifies nothing on the five models.
We exclude ERAN's CPU DeepPoly,
which is orders of magnitude slower.

\subsection{RQ1: Validation on Known Counterexamples}

The instances are the certified points~$x$ of the counterexamples of \cref{sec:cex}, on the
three image models of that section: for the naturally trained models, those of the $30$
counterexamples per model and per format of the continuous search, at float32, float64 and
float16, at the radii \cref{sec:cex} reports; for the adversarially biased variants, the $24$, $204$ and
$370$ certified test images whose certificates the grid search refutes at the standard radii.
We ask whether our certifier certifies~$x$, for the execution the counterexample was found
against, at the radius at which the real-arithmetic certifier does, using the tighter
per-model Gram counts of \cref{app:training}.

\paragraph{Finding} Our certifier rejects all instances, on all three models, formats
and modes (CIFAR-10 at float16 by refusal since $n_1 u = 1.5$); each certificate is refuted under the
execution it was issued for, so rejection is the only sound answer. On the biased models it certifies nothing at the standard~<radii.

\subsection{RQ2: The Cost of Floating-Point Soundness, and Reclaiming It Pre-Deployment}\label{sec:rq2}

We measure the floating-point cost of \cref{sec:eval-setup} on the full test set of every
model, how much of it the pre-deployment modes of \cref{sec:hybrid} reclaim, the time to
certify, and how cost and time scale with network width. The cost is the gap between the
\emph{Real} and \emph{Std} columns of the Robustness block of \cref{tab:master}, which reports
the certified rate and so isolates the certifier's conservatism from model accuracy; what the
pre-deployment modes reclaim is measured on VRA, the quantity of interest pre-deployment
(\cref{sec:eval-setup}); the times are those of \cref{tab:timing}.

\paragraph{Finding: the cost is small across task families} Standard
mode certifies a substantial fraction on benchmarks spanning all task families;
the floating-point cost
is small on five of six models (\cref{tab:master}).

\paragraph{Finding: CIFAR-10 is the exception, and not because of the network's scale}
CIFAR-10's cost is $27.2$\,pp; however, EMNIST-ByClass has the
same architecture and costs only $1.16$\,pp. What drives the cost is the certification task, captured
by the input radius relative to the certified perturbation, $r_0/\varepsilon$. Define the \emph{conservatism ratio} as
$\mathbb{E}_{x,j}\big[E_{\text{ctr}}^{(j,i^*)} + E_{\text{ball}}^{(j,i^*)}\big] \big/
\mathbb{E}_{x,j}\big[\varepsilon\, L_{j,i^*}\big]$, the mean floating-point margin degradation
of the certification check (\cref{thm:fp-robust}) divided by the mean real-arithmetic
certification margin, both means taken over all test points~$x$ and all competing
classes~$j \ne i^*$.
Across the three models that isolate this axis it climbs with $r_0/\varepsilon$: EMNIST-ByClass
($r_0/\varepsilon = 35$) has ratio $0.08$, HIGGS-512 ($52$) has $0.17$, and CIFAR-10 ($207$) has
$1.09$, some $14\times$ EMNIST-ByClass's, because CIFAR-10's high-dimensional inputs give
a far larger $r_0$ ($29.2$ against~$10.6$). The task is hard before floating point enters:
adversarial attacks on the float32 execution (\cref{app:training}) show that at least $33.4\%$ of
CIFAR-10's test points are not robust. That leaves $20.5\%$ of the test
set neither certified by the real-arithmetic check nor refuted, which suggests that the global
Lipschitz bound is loose for CIFAR-10.

\paragraph{Finding: the pre-deployment modes reclaim most of the conservatism, where there is any
to reclaim} In \cref{tab:master}'s VRA columns, Hybrid-Centre and then Measured-Radii narrow
the gap to the real-arithmetic ceiling, most where the Standard mode is most
conservative: on CIFAR-10 and EMNIST-Balanced Measured-Radii closes over half the gap; on HIGGS it closes
four-fifths. Where the Standard mode is near-tight (MNIST, Fashion MNIST, EMNIST-ByClass), little
remains to reclaim.

\paragraph{Finding: certification is fast} The one-off spectral-norm computation takes seconds
to a few minutes per model (\cref{tab:timing}): CIFAR-10's takes $19.5$\,s, against tens of
hours for the exact-rational computation~\cite{tobler2025}, whose bounds it matches to within
$3.5\times10^{-11}$ relative per layer on the three image models. Per-instance certification
then takes a few milliseconds in every mode, and is nearly insensitive to the number of classes,
since the certifier's work per class is linear: the $62$-class EMNIST-ByClass takes
less time than the two-class HIGGS-1024.

\paragraph{Finding: with width, the cost and the time grow modestly} The HIGGS width sweep of
\cref{sec:eval-setup} holds the task fixed and varies only the hidden width, from $128$ to
$1024$. Across it the real-arithmetic certified rate holds near $82.5\%$, so model capacity is
not the bottleneck, while the Standard-mode floating-point cost climbs from $0.42$ to
$5.93$\,pp; the Measured-Radii mode reclaims $77$--$83\%$ of it, holding the cost to at most
$1.4$\,pp. The one-off norm computation grows roughly quadratically with width, about $4\times$
per doubling, from $2.8$ to $192$\,s, and per-instance certification time in the Standard mode
from $0.5$ to $4.1$\,ms. Added width thus raises the cost modestly and recoverably.

\begin{table}[t]
\centering
\small
\setlength{\tabcolsep}{4pt}
\begin{tabular}{l r rrr rr}
\toprule
 & & \multicolumn{3}{c}{Ours (ms/inst)} & \multicolumn{2}{c}{GPUPoly (ms/inst)} \\
\cmidrule(lr){3-5}\cmidrule(lr){6-7}
Model & Norms (s) & Std & Hyb & Meas & comparator & level~3 \\
\midrule
MNIST           & 5.5   & 1.95  & 3.21  & 5.30  &  92 & 104 \\
Fashion MNIST   & 10.3  & 3.82  & 5.66  & 8.96  & 204 & 231 \\
CIFAR-10        & 19.5  & 5.68  & 9.92  & 14.27 & 172 & 195 \\
HIGGS-1024      & 192.4 & 4.06  & 8.87  & 13.87 & --  & --  \\
EMNIST-ByClass  & 18.4  & 3.52  & 5.27  & 8.03  & 142 & 163 \\
EMNIST-Balanced & 90.8  & 7.65  & 11.09 & 16.35 & 547 & 631 \\
\bottomrule
\end{tabular}
\caption{\label{tab:timing}%
  Execution time (12 Gram iterations). \emph{Norms} is our one-off, per-model spectral-norm
  computation, which the per-instance times exclude and for which GPUPoly has no counterpart.
  Per-instance times are means: ours on the laptop CPU of \cref{sec:eval-setup} in the three
  modes, GPUPoly's on the A10 GPU for the comparator at float32 and its shipped float64
  configuration, level~3 (\textbf{RQ3}), as ERAN's front end reports them, over all test points. HIGGS has no GPUPoly
  column (\cref{sec:eval-setup}).}
\end{table}

\subsection{RQ3: Comparison with ERAN}\label{sec:rq3}

\Cref{tab:eran} sets the VRA of every mode beside DeepPoly's, as GPUPoly computes it on the
full test set of every image model as configured in \cref{sec:eval-setup};
\cref{tab:timing} the per-instance times.

\begin{table}[t]
\centering
\footnotesize
\setlength{\tabcolsep}{3pt}
\begin{tabular}{l c cccc ccc cc}
\toprule
 & & \multicolumn{4}{c}{Ours: VRA (\%)} & \multicolumn{3}{c}{DeepPoly (GPUPoly): VRA (\%)} & \multicolumn{2}{c}{Retained (\%)} \\
\cmidrule(lr){3-6}\cmidrule(lr){7-9}\cmidrule(lr){10-11}
 & & & \multicolumn{3}{c}{float32} & & \multicolumn{2}{c}{float32} & & \\
\cmidrule(lr){4-6}\cmidrule(lr){8-9}
Model & $\varepsilon_{\ell_\infty}$ & Real & Std & Hyb & Meas & Real & comparator & level~4 & Meas & DeepPoly \\
\midrule
MNIST           & 0.0107  & 95.40 & 95.17 & 95.28 & 95.31 & 97.09 & 96.73 & 90.46 & 99.9 & 99.6 \\
Fashion MNIST   & 0.00892 & 79.54 & 78.47 & 78.94 & 79.08 & 49.43 & 17.49 &  0.02 & 99.4 & 35.4 \\
CIFAR-10        & 0.00254 & 35.95 & 16.96 & 25.57 & 27.52 & 41.26 &  3.00 &  0.02 & 76.6 &  7.3 \\
EMNIST-ByClass  & 0.0107  & 75.37 & 74.54 & 74.97 & 75.01 & 74.46 & 52.13 &  0.00 & 99.5 & 70.0 \\
EMNIST-Balanced & 0.0107  & 71.25 & 68.33 & 69.74 & 70.06 & 54.54 &  0.69 &  0.00 & 98.3 &  1.3 \\
\bottomrule
\end{tabular}
\caption{\label{tab:eran}%
  Comparison with ERAN's GPUPoly~\cite{GPUPoly} on each model's full test set, for the float32
  round-to-nearest execution. \emph{Real}, \emph{Std},
  \emph{Hyb} and \emph{Meas} are as in \cref{tab:master}. For DeepPoly, \emph{Real} is its real-arithmetic
  ceiling (\cref{sec:eval-setup}), whose results coincide with its shipped float64 configuration, level~3 (\cref{app:gpupoly});
  \emph{Level~4} is GPUPoly's float32 configuration; the generous \emph{comparator} under-charges its
  float32 rounding (\cref{sec:eval-setup}). At float64 both tools
  certify exactly their real-arithmetic set. \emph{Retained} is the float32 VRA as a percentage of the tool's
  own real-arithmetic VRA (ours in the Measured-Radii mode). GPUPoly's $\ell_\infty$ radius is
  derived from ours as in \cref{sec:eval-setup}, shown rounded down to three significant figures.}
\end{table}

\paragraph{Finding: at float64, floating-point soundness is free for both tools}
At float64 neither tool pays for floating-point soundness: every one of our modes certifies
exactly the real-arithmetic set
(the \emph{Real} columns of \cref{tab:eran})
on every model, as does GPUPoly's shipped float64 analysis.
The accounting is still needed at float64, as the float64 counterexamples of
\cref{sec:cex} demonstrate.
The difference between the two tools' real-arithmetic VRA is the difference in precision between
our global Lipschitz bound and DeepPoly's per-neuron relaxation. On its smaller $\ell_\infty$
box, DeepPoly verifies more than we do on CIFAR-10, less on Fashion MNIST and
EMNIST-Balanced, and about the same on MNIST and EMNIST-ByClass.

\paragraph{Finding: at float32, ours retains most of its real-arithmetic VRA; DeepPoly
does not} DeepPoly's float32 numbers are the comparator's, which under-charges GPUPoly's
float32 rounding (\cref{sec:eval-setup}). Our Measured-Radii mode retains $98.3\%$ or more of its real-arithmetic VRA on four models and
$76.6\%$ on CIFAR-10; DeepPoly retains $99.6\%$ on MNIST but between $1.3\%$ and $70.0\%$ on
the other four. This difference is caused by the different ways the two tools account for rounding:
our error term is a single additive quantity on the output margin that compounds
per layer via the layer spectral norms; DeepPoly's rounding
charges attach to each product of the analysis and compound through the depth-times-width
structure that makes its abstraction precise. In absolute terms, ours reaches a higher VRA than DeepPoly on four models
in every mode, and is within $1.6$\,pp on MNIST.

\paragraph{Finding: on CIFAR-10, DeepPoly's floating-point soundness hands back its
abstraction's advantage} CIFAR-10 is the one model on which DeepPoly's abstraction verifies more
than our global Lipschitz bound in real arithmetic ($41.26\%$ against $35.95\%$, on its smaller
box), the model on which \cref{sec:rq2} finds that bound substantially loose. Once rounding is
charged the order reverses: ours loses $8.4$\,pp of VRA in the Measured-Radii mode, DeepPoly
$38.3$\,pp, over nine tenths of its ceiling.

\paragraph{Finding: ours is an order of magnitude faster, or more}
Our certifier takes $2$--$16$\,ms per instance on the laptop CPU in every mode, against
$92$--$547$\,ms for the GPUPoly comparator (as reported by ERAN) on the A10 GPU and $104$--$631$\,ms for its level~3
(\cref{tab:timing}).
These per-instance times exclude
our one-off spectral-norm computation ($5$--$91$\,s per image model);
amortised over the test set it adds at most $4.8$\,ms per instance
and leaves the Measured-Radii mode $11$--$26\times$ faster.

\section{Related Work}

Our treatment of rounding in the network's execution and in the certifier's own computation,
the two places of \cref{sec:intro}, rests on the classical error analysis of linear
algebra~\cite{wilkinson1960error,demmel1997applied,higham2002}, whose key results are
mechanised in the LAProof library~\cite{LAProof}, on which our Rocq formalisation builds.

\paragraph{Rounding in the network's execution}
Jia and Rinard~\cite{Jia_Rinard_SAS21} first showed that it yields false certificates from
\emph{complete} robustness verifiers, by searching for inputs that sit almost
exactly~$\varepsilon$ from the decision boundary. Zombori et al.~\cite{zombori2021fooling}
construct backdoored networks with trigger inputs for which a complete verifier claims robustness
under its arithmetic model. Sz{\'a}sz et al.~\cite{szasz2025no} plant backdoors whose trigger is
not an input but the execution environment itself, such as its summation order or arithmetic
precision. None concerns Lipschitz-based certification, where it was open whether the method's
conservatism absorbs the rounding. Our counterexamples (\cref{sec:cex}) show that it does not,
against a certifier that is moreover formally verified. Runtime details that depart from standard
floating-point semantics remain outside our theory (\cref{sec:conclusion}).

Beuzeville et al.~\cite{beuzeville2026deterministic} and El Arar et al.~\cite{elarar2025mixed}
apply classical error analysis to network inference: Beuzeville et al.\ give a componentwise
backward-error result (the computed output is the exact output of a slightly perturbed network),
El Arar et al.\ a forward componentwise analysis that guides mixed-precision accumulation, a
higher-precision execution of the numerically sensitive inner products. Our pre-deployment modes
(\cref{sec:hybrid}) also run a higher-precision execution, but to \emph{measure} floating-point
deviation, which the bound then carries over the ball. Both their analyses bound the rounding of the one execution at the input given, and both
assume that no overflow occurs. Our certificates instead rest on a forward normwise deviation
bound (\cref{cor:recursion}) and the overflow checks of \cref{sec:overflow}, both of which
cover every input in the $\varepsilon$-ball. Their analyses handle rounding in the activations
and could extend ours beyond FP-exact ones (assumption~\ref{assumption-fp-exact}).
Differential verification~\cite{reludiff,fanc,compaqt} also bounds a discrepancy over an input region, between a
network and its pruned or quantised version, with both assumed to execute in real arithmetic.

Verifiers sound for the network's floating-point execution exist. DeepZ~\cite{DeepZ} and
DeepPoly~\cite{DeepPoly}, in ERAN~\cite{ERAN}, are based on abstract interpretation;
FMIPVerify~\cite{FMIPVerify} on mixed-integer linear programming over a real-interval
abstraction; and QNNVerifier~\cite{song2021qnnverifier} and Manino et
al.~\cite{manino2025floating} translate the network into C for floating-point-sound program
verifiers. Each pays for the accounting. At float32 GPUPoly's single-precision analysis is
near-vacuous on four of our five models, and even with the rounding charged in a
double-precision analysis it loses most of its real-arithmetic certificates on three
(\cref{sec:rq3}). FMIPVerify is evaluated on downsampled ($7 \times 7$) MNIST and Fashion MNIST
models an order of magnitude smaller than ours, and the program verifiers that the translations
target remain limited to small networks: Manino et al.'s benchmark~\cite{manino2025floating}
reaches 170K parameters, below our smallest model's 216K, and the verifiers solve few of its
instances.

Lipschitz-based certification rose to prominence with Leino et al.~\cite{leino2021} and now
scales to billion-parameter models~\cite{hu2026lipnext}. Its constants are \emph{global}, as ours
are, or \emph{local} to a region around the input and tighter: DeepJ~\cite{DeepJ} bounds local
constants soundly through a dual-number abstraction, and Laurel et
al.~\cite{laurel2023synthesizing} synthesise more precise such analysers. DeepJ can moreover
add the per-operation rounding terms that DeepPoly uses. That option yields no certification
check sound for the floating-point execution, which is not a Lipschitz function (\cref{sec:intro}). Our check
(\cref{thm:fp-robust}) utilises a real-arithmetic constant that it requires
to be valid on the ball (condition~(ii)), so a sound local constant could suffice. We use
global constants because they are computed once per model, after which certifying an input takes
milliseconds; local constants are recomputed per input.

\paragraph{Rounding in the certifier's own computation}
Jin et al.~\cite{jin2024around} show that it inflates computed certified radii past their
real-arithmetic values and find adversarial examples inside the inflation. Rounded interval
arithmetic, as in PyRAT~\cite{pyrat2024}, is an established remedy.
Tobler et al.~\cite{tobler2025} make the certifier's computation exact, by rational arithmetic
in a certifier that is itself verified, at the price of spectral norms prohibitively slow to
compute at scale. We keep rational arithmetic everywhere but the spectral norms
(\cref{sec:eval-setup}), made feasible by our sound float64 Gram iteration
(\cref{sec:norms}).

\paragraph{Real arithmetic assumed throughout}
The remaining robustness verifiers reason symbolically, often over much larger networks than ours~\cite{VNNCOMP2025}.
Among them are
$\alpha,\beta$-CROWN~\cite{xu2020automatic,xu2021fast,wang2021betacrown,zhang2022general,shi2024genbab},
Marabou~\cite{katz2019marabou,wu2024marabou},
NeuralSAT~\cite{duong2023dpllt,duong2024harnessing,duong2025neuralsat},
nnenum~\cite{bak2020cav,bak2021nnenum},
CORA~\cite{althoff_2015,koller_et_al_2025settraining,koller_et_al_2025shadows,kochdumper_et_al_2023,ladner_althoff_2023},
NeVer2~\cite{demarchi2024never2,guidotti2021pynever} and NNV~\cite{tran2020cav_tool,manzanas2023cav}.
They assume real arithmetic in the network's execution and their own computation.

\section{Conclusion}\label{sec:conclusion}

We presented the first Lipschitz-based robustness certification method that is both efficient
and sound with respect to floating-point execution, motivated by rounding-induced
counterexamples that refute the certificates of a formally verified real-arithmetic certifier.
Our certification check subtracts from the margin a
bound on the deviation between floating-point and real-arithmetic execution, under checked
overflow conditions; measurements of a high-precision execution tighten the bound
pre-deployment; and Gram iteration in float64 with its rounding tracked bounds the spectral norms
soundly, orders of magnitude faster than in exact rationals. The theory is mechanised in Rocq.
Our certifier certifies the full test sets of image, tabular and many-class models at
milliseconds per instance, losing little verified robust accuracy except on CIFAR-10, where the
cost is set by the task rather than by the network's size and the pre-deployment modes reclaim
much of it (\cref{sec:rq2}). Compared against ERAN at float32 (\cref{sec:rq3}), it retains most of
its real-arithmetic verified robust accuracy where GPUPoly's per-product rounding terms
lose most of theirs on three of the five models (including CIFAR-10), in more than an order of magnitude less
time per instance; to our knowledge it is the first floating-point-sound verifier that yields
useful certificates at float32 on every network of this size that we tried. It is sound in every regime we tested, though vacuous for the
adversarially constructed models and at float16, where DeepPoly's accounting verifies
nothing either (\cref{sec:eval-setup}).

Our certifier and evaluation are specialised to dense ReLU networks with identity output
activation; the theory is more general. The layer lemmas and \cref{thm:fp-robust} already
hold for arbitrary weight matrices and any activation satisfying the assumptions of
\cref{sec:nn-model} (e.g., identity, ReLU, absolute value,
hard tanh), so a zero-padded convolution, a linear operator, is likely covered already,
given sound upper bounds on its spectral norms~\cite{delattre2024,grishina2024}.
Flush-to-zero execution changes only the per-operation constants (see \cref{app:ftz}, which is
not mechanised). Activations that round, unfolded batch normalisation and residual connections each introduce
rounding at a point the theory does not charge; we expect each to add one term to a
bound already proved, and existing analyses~\cite{beuzeville2026deterministic,elarar2025mixed}
show how the first could be charged. Future work
lies in two directions. Firstly, supporting an $\ell_\infty$ certificate by replacing the
spectral norms by $\ell_\infty$ operator norms. Secondly, reasoning about fused operators, hardware-specific kernels and
reduced-precision accelerators, whose mismatch with an assumed semantics undermines any
verifier~\cite{szasz2025no}. Modelling accelerators' wide accumulation of float16 products
in float32, whose error analysis is established~\cite{blanchard2020mixed,elarar2025mixed},
would bring sound certification to the precisions where rounding is largest.

% Data-Availability Statement removed for the anonymous POPL submission: POPL
% requires no DAS, and the artifact will go via the submission system's
% supplementary material (Coq proofs + README, plus a link to the full artifact
% for reviewers who want to build the proofs). Re-add a DAS for camera-ready.

\bibliographystyle{ACM-Reference-Format}
\bibliography{references}

@inproceedings{jin2024around,
  title={Getting a-Round Guarantees: Floating-Point Attacks on Certified Robustness},
  author={Jin, Jiankai and Ohrimenko, Olga and Rubinstein, Benjamin I. P.},
  booktitle={Proceedings of the 2024 Workshop on Artificial Intelligence and Security (AISec)},
  year={2024},
  doi={10.1145/3689932.3694761}
}

@inproceedings{reluplex,
  title={Reluplex: An efficient {SMT} solver for verifying deep neural networks},
  author={Katz, Guy and Barrett, Clark and Dill, David L and Julian, Kyle and Kochenderfer, Mykel J},
  booktitle={International Conference on Computer Aided Verification (CAV)},
  pages={97--117},
  year={2017},
  organization={Springer}
}

@article{DeepZ,
  title={Fast and effective robustness certification},
  author={Singh, Gagandeep and Gehr, Timon and Mirman, Matthew and P{\"u}schel, Markus and Vechev, Martin},
  journal={Advances in neural information processing systems},
  volume={31},
  year={2018}
}

@inproceedings{leino2021,
  author       = {Klas Leino and
                  Zifan Wang and
                  Matt Fredrikson},
  title        = {Globally-Robust Neural Networks},
  booktitle    = {International Conference on Machine Learning (ICML)},
  series       = {Proceedings of Machine Learning Research},
  volume       = {139},
  pages        = {6212--6222},
  publisher    = {{PMLR}},
  year         = {2021},
  url          = {http://proceedings.mlr.press/v139/leino21a.html},
  bibsource    = {dblp computer science bibliography, https://dblp.org}
}

@inproceedings{weng2018towards,
  title={Towards fast computation of certified robustness for relu networks},
  author={Weng, Lily and Zhang, Huan and Chen, Hongge and Song, Zhao and Hsieh, Cho-Jui and Daniel, Luca and Boning, Duane and Dhillon, Inderjit},
  booktitle={International Conference on Machine Learning (ICML)},
  pages={5276--5285},
  year={2018},
  organization={PMLR}
}

@inproceedings{tobler2025,
  title={A Formally Verified Robustness Certifier for Neural Networks},
  author={Tobler, James and Syeda, Hira Taqdees and Murray, Toby},
  booktitle={International Conference on Computer Aided Verification (CAV)},
  pages={327--348},
  year={2025},
  organization={Springer}
}

@article{song2021qnnverifier,
  title={{QNNVerifier}: A tool for verifying neural networks using {SMT}-based model checking},
  author={Song, Xidan and Manino, Edoardo and Sena, Luiz and Alves, Erickson and de Lima Filho, Eddie and Bessa, Iury and Lujan, Mikel and Cordeiro, Lucas},
  journal={arXiv preprint arXiv:2111.13110},
  year={2021}
}

@inproceedings{FMIPVerify,
  title={Sound Floating-Point Neural Network Verification with {MILP}},
  author={Yang, Shifu and Chen, Liqian and Yin, BangHu and Li, MingHao and Zhou, Yuan and Wang, Ji},
  booktitle={Asia-Pacific Software Engineering Conference (APSEC)},
  pages={01--10},
  year={2024},
  organization={IEEE}
}

@inproceedings{Jia_Rinard_SAS21,
  title={Exploiting verified neural networks via floating point numerical error},
  author={Jia, Kai and Rinard, Martin},
  booktitle={International Static Analysis Symposium},
  pages={191--205},
  year={2021},
  organization={Springer}
}

@article{gouk2021,
  author       = {Henry Gouk and
                  Eibe Frank and
                  Bernhard Pfahringer and
                  Michael J. Cree},
  title        = {Regularisation of neural networks by enforcing {Lipschitz} continuity},
  journal      = {Mach. Learn.},
  volume       = {110},
  number       = {2},
  pages        = {393--416},
  year         = {2021},
  url          = {https://doi.org/10.1007/s10994-020-05929-w},
  doi          = {10.1007/S10994-020-05929-W},
  bibsource    = {dblp computer science bibliography, https://dblp.org}
}

@inproceedings{delattre2023,
  title={Efficient bound of {Lipschitz} constant for convolutional layers by {Gram} iteration},
  author={Delattre, Blaise and Barth{\'e}lemy, Quentin and Araujo, Alexandre and Allauzen, Alexandre},
  booktitle={International Conference on Machine Learning (ICML)},
  pages={7513--7532},
  year={2023},
  organization={PMLR}
}

@article{delattre2024,
  title={Spectral norm of convolutional layers with circular and zero paddings},
  author={Delattre, Blaise and Barth{\'e}lemy, Quentin and Allauzen, Alexandre},
  journal={arXiv preprint arXiv:2402.00240},
  year={2024}
}

@inproceedings{grishina2024,
  title={Tight and efficient upper bound on spectral norm of convolutional layers},
  author={Grishina, Ekaterina and Gorbunov, Mikhail and Rakhuba, Maxim},
  booktitle={European Conference on Computer Vision},
  pages={19--34},
  year={2024},
  organization={Springer}
}

@inproceedings{reludiff,
  title={{ReluDiff}: Differential Verification of Deep Neural Networks},
  author={Paulsen, Brandon and Wang, Jingbo and Wang, Chao},
  booktitle={International Conference on Software Engineering (ICSE)},
  pages={714--726},
  year={2020},
  organization={IEEE/ACM}
}

@article{fanc,
  title={Proof Transfer for Fast Certification of Multiple Approximate Neural Networks},
  author={Ugare, Shubham and Singh, Gagandeep and Misailovic, Sasa},
  journal={Proceedings of the ACM on Programming Languages},
  volume={6},
  number={OOPSLA1},
  year={2022},
  doi={10.1145/3527319}
}

@article{compaqt,
  title={Quantization with Guaranteed Floating-Point Neural Network Classifications},
  author={Kabaha, Anan and Drachsler-Cohen, Dana},
  journal={Proceedings of the ACM on Programming Languages},
  volume={9},
  number={OOPSLA2},
  year={2025},
  doi={10.1145/3763118}
}

@inproceedings{madry2018,
  title={Towards Deep Learning Models Resistant to Adversarial Attacks},
  author={Madry, Aleksander and Makelov, Aleksandar and Schmidt, Ludwig and Tsipras, Dimitris and Vladu, Adrian},
  booktitle={International Conference on Learning Representations (ICLR)},
  year={2018}
}

@inproceedings{DeepFool,
  title={{DeepFool}: a simple and accurate method to fool deep neural networks},
  author={Moosavi-Dezfooli, Seyed-Mohsen and Fawzi, Alhussein and Frossard, Pascal},
  booktitle={IEEE Conference on Computer Vision and Pattern Recognition (CVPR)},
  pages={2574--2582},
  year={2016}
}

@article{ART,
  title={{Adversarial Robustness Toolbox} v1.0.0},
  author={Nicolae, Maria-Irina and Sinn, Mathieu and Tran, Minh Ngoc and Buesser, Beat and Rawat, Ambrish and Wistuba, Martin and Zantedeschi, Valentina and Baracaldo, Nathalie and Chen, Bryant and Ludwig, Heiko and others},
  journal={arXiv preprint arXiv:1807.01069},
  year={2018}
}

@techreport{CIFAR10,
  title={Learning Multiple Layers of Features from Tiny Images},
  author={Alex Krizhevsky},
  institution={University of Toronto},
  year={2009}
}

@article{MNIST,
  title={Gradient-based learning applied to document recognition},
  author={LeCun, Yann and Bottou, L{\'e}on and Bengio, Yoshua and Haffner, Patrick},
  journal={Proceedings of the IEEE},
  volume={86},
  number={11},
  pages={2278--2324},
  year={1998},
  publisher={IEEE}
}

@article{Fashion_MNIST,
  title={Fashion-{MNIST}: a novel image dataset for benchmarking machine learning algorithms},
  author={Xiao, Han and Rasul, Kashif and Vollgraf, Roland},
  journal={arXiv preprint arXiv:1708.07747},
  year={2017}
}

@article{baldi2014,
  title={Searching for Exotic Particles in High-Energy Physics with Deep Learning},
  author={Baldi, Pierre and Sadowski, Peter and Whiteson, Daniel},
  journal={Nature Communications},
  volume={5},
  pages={4308},
  year={2014},
  doi={10.1038/ncomms5308}
}

@inproceedings{cohen2017emnist,
  title={{EMNIST}: Extending {MNIST} to Handwritten Letters},
  author={Cohen, Gregory and Afshar, Saeed and Tapson, Jonathan and van Schaik, Andr\'{e}},
  booktitle={International Joint Conference on Neural Networks (IJCNN)},
  pages={2921--2926},
  year={2017},
  doi={10.1109/IJCNN.2017.7966217}
}

@book{higham2002,
  title={Accuracy and stability of numerical algorithms},
  author={Higham, Nicholas J},
  year={2002},
  publisher={SIAM}
}

@inproceedings{LAProof,
  title={{LAProof}: A library of formal proofs of accuracy and correctness for linear algebra programs},
  author={Kellison, Ariel E and Appel, Andrew W and Tekriwal, Mohit and Bindel, David},
  booktitle={2023 IEEE 30th Symposium on Computer Arithmetic (ARITH)},
  pages={36--43},
  year={2023},
  organization={IEEE}
}

@article{DeepPoly,
  title={An abstract domain for certifying neural networks},
  author={Singh, Gagandeep and Gehr, Timon and P{\"u}schel, Markus and Vechev, Martin},
  journal={Proceedings of the ACM on Programming Languages},
  volume={3},
  number={POPL},
  pages={1--30},
  year={2019},
  publisher={ACM New York, NY, USA}
}

@article{DeepJ,
  title   = {A Dual Number Abstraction for Static Analysis of {Clarke} {Jacobians}},
  author  = {Laurel, Jacob and Yang, Rem and Singh, Gagandeep and Misailovic, Sasa},
  journal = {Proceedings of the ACM on Programming Languages},
  volume  = {6},
  number  = {POPL},
  pages   = {1--30},
  year    = {2022},
  doi     = {10.1145/3498718}
}

@article{laurel2023synthesizing,
  title   = {Synthesizing Precise Static Analyzers for Automatic Differentiation},
  author  = {Laurel, Jacob and Qian, Siyuan Brant and Singh, Gagandeep and Misailovic, Sasa},
  journal = {Proceedings of the ACM on Programming Languages},
  volume  = {7},
  number  = {OOPSLA2},
  pages   = {1964--1992},
  year    = {2023},
  doi     = {10.1145/3622867}
}

@inproceedings{croce2020autoattack,
  title     = {Reliable Evaluation of Adversarial Robustness with an Ensemble of Diverse Parameter-free Attacks},
  author    = {Croce, Francesco and Hein, Matthias},
  booktitle = {Proceedings of the 37th International Conference on Machine Learning ({ICML})},
  year      = {2020}
}

@inproceedings{GPUPoly,
  title     = {Scaling Polyhedral Neural Network Verification on {GPUs}},
  author    = {M{\"u}ller, Christoph and Serre, Fran{\c{c}}ois and Singh, Gagandeep and P{\"u}schel, Markus and Vechev, Martin},
  booktitle = {Proceedings of Machine Learning and Systems ({MLSys})},
  volume    = {3},
  year      = {2021}
}

@misc{ERAN,
    title        = {{ERAN}: {ETH} Robustness Analyzer for Neural Networks},
    author       = {M{\"u}ller, Mark Niklas and
                    Singh, Gagandeep and
                    Balunovic, Mislav and
                    Makarchuk, Gleb and
                    Ruoss, Anian and
                    Serre, Fran{\c{c}}ois and
                    Baader, Maximilian and
                    Drachsler-Cohen, Dana and
                    Gehr, Timon and
                    Hoffmann, Adrian and
                    Maurer, Jonathan and
                    M{\"u}ller, Christoph and
                    P{\"u}schel, Markus and
                    Tsankov, Petar and
                    Vechev, Martin},
    howpublished = {\url{https://github.com/eth-sri/eran}},
    year         = {2018},
    note         = {Secure, Reliable, and Intelligent Systems Lab, ETH Z{\"u}rich}
  }

@article{VNNCOMP2025,
  title={The 6th International Verification of Neural Networks Competition ({VNN-COMP 2025}): Summary and Results},
  author={Kaulen, Konstantin and Ladner, Tobias and Bak, Stanley and Brix, Christopher and Duong, Hai and Flinkow, Thomas and Johnson, Taylor T and Koller, Lukas and Manino, Edoardo and Nguyen, ThanhVu H and others},
  journal={arXiv preprint arXiv:2512.19007},
  year={2025}
}

@article{xu2020automatic,
  title={Automatic perturbation analysis for scalable certified robustness and beyond},
  author={Xu, Kaidi and Shi, Zhouxing and Zhang, Huan and Wang, Yihan and Chang, Kai-Wei and Huang, Minlie and Kailkhura, Bhavya and Lin, Xue and Hsieh, Cho-Jui},
  journal={Advances in Neural Information Processing Systems},
  volume={33},
  year={2020}
}

@inproceedings{xu2021fast,
    title={{Fast and Complete}: Enabling Complete Neural Network Verification with Rapid and Massively Parallel Incomplete Verifiers},
    author={Kaidi Xu and Huan Zhang and Shiqi Wang and Yihan Wang and Suman Jana and Xue Lin and Cho-Jui Hsieh},
    booktitle={International Conference on Learning Representations},
    year={2021},
    url={https://openreview.net/forum?id=nVZtXBI6LNn}
}

@article{wang2021betacrown,
  title={{Beta-CROWN}: Efficient Bound Propagation with Per-neuron Split Constraints for Complete and Incomplete Neural Network Verification},
  author={Wang, Shiqi and Zhang, Huan and Xu, Kaidi and Lin, Xue and Jana, Suman and Hsieh, Cho-Jui and Kolter, Zico},
  journal={arXiv preprint arXiv:2103.06624},
  year={2021}
}

@article{zhang2022general,
  title={General cutting planes for bound-propagation-based neural network verification},
  author={Zhang, Huan and Wang, Shiqi and Xu, Kaidi and Li, Linyi and Li, Bo and Jana, Suman and Hsieh, Cho-Jui and Kolter, J Zico},
  journal={Advances in Neural Information Processing Systems (NeurIPS)},
  year={2022}
}

@article{shi2024genbab,
  title={Neural Network Verification with Branch-and-Bound for General Nonlinearities},
  author={Shi, Zhouxing and Jin, Qirui and Kolter, Zico and Jana, Suman and Hsieh, Cho-Jui and Zhang, Huan},
  journal={arXiv preprint arXiv:2405.21063},
  year={2024}
}

@article{pyrat2024,
    title={Neural Network Verification with {PyRAT}},
        author={Lemesle, Augustin and Lehmann, Julien and Le Gall, Tristan},
        journal={arXiv preprint arXiv:2410.23903},
        year={2024}
}

@inproceedings{katz2019marabou,
  title={The {Marabou} framework for verification and analysis of deep neural networks},
  author={Katz, Guy and Huang, Derek A and Ibeling, Duligur and Julian, Kyle and Lazarus, Christopher and Lim, Rachel and Shah, Parth and Thakoor, Shantanu and Wu, Haoze and Zelji{\'c}, Aleksandar and others},
  booktitle={International Conference on Computer Aided Verification},
  pages={443--452},
  year={2019},
  organization={Springer}
}

@inproceedings{wu2024marabou,
  title={Marabou 2.0: a versatile formal analyzer of neural networks},
  author={Wu, Haoze and Isac, Omri and Zelji{\'c}, Aleksandar and Tagomori, Teruhiro and Daggitt, Matthew and Kokke, Wen and Refaeli, Idan and Amir, Guy and Julian, Kyle and Bassan, Shahaf and others},
  booktitle={International Conference on Computer Aided Verification},
  pages={249--264},
  year={2024},
  organization={Springer}
}

@inproceedings{duong2025neuralsat,
  title={{NeuralSAT}: A High-Performance Verification Tool for Deep Neural Networks},
  author={Duong, Hai and Nguyen, ThanhVu and Dwyer, Matthew B},
  booktitle={International Conference on Computer Aided Verification},
  pages={409--423},
  year={2025},
  organization={Springer}
}

@article{duong2024harnessing,
  title={Harnessing neuron stability to improve {DNN} verification},
  author={Duong, Hai and Xu, Dong and Nguyen, ThanhVu and Dwyer, Matthew B},
  journal={Proceedings of the ACM on Software Engineering},
  volume={1},
  number={FSE},
  pages={859--881},
  year={2024},
  publisher={ACM New York, NY, USA}
}

@misc{duong2023dpllt,
      title={A {DPLL(T)} Framework for Verifying Deep Neural Networks},
      author={Hai Duong and Linhan Li and ThanhVu Nguyen and Matthew Dwyer},
      year={2023},
      note={arXiv, 25 pages},
      eprint={2307.10266},
      archivePrefix={arXiv},
      primaryClass={cs.LG}
}

@inproceedings{bak2020cav,
author = "Stanley Bak and Hoang-Dung Tran and Kerianne Hobbs and Taylor T. Johnson",
title = "Improved Geometric Path Enumeration for Verifying {ReLU} Neural Networks",
booktitle = "32nd International Conference on Computer-Aided Verification (CAV)",
year = "2020",
month = "July",
}

@inproceedings{bak2021nnenum,
  title={nnenum: Verification of {ReLU} Neural Networks with Optimized Abstraction Refinement},
  author={Bak, Stanley},
  booktitle={NASA Formal Methods Symposium},
  pages={19--36},
  year={2021},
  organization={Springer}
}

@article{koller_et_al_2025settraining,
  title={Set-Based Training for Neural Network Verification},
  author={Koller, Lukas and Ladner, Tobias and Althoff, Matthias},
  journal={TMLR},
  year={2025}
}

@article{koller_et_al_2025shadows,
    title={Out of the Shadows: Exploring a Latent Space for Neural Network Verification},
    author={Lukas Koller and Tobias Ladner and Matthias Althoff},
    journal={arXiv},
    year={2025}
}

@inproceedings{althoff_2015,
  author    = {Althoff, Matthias},
  booktitle = {Proc. of the Workshop on Applied Verification for Continuous and Hybrid Systems (ARCH)},
  pages     = {120--151},
  title     = {An Introduction to {CORA} 2015},
  year      = {2015}
}

@inproceedings{kochdumper_et_al_2023,
  author    = {Kochdumper, Niklas and Schilling, Christian and Althoff, Matthias and Bak, Stanley},
  booktitle = {NASA Formal Methods},
  pages     = {16--36},
  title     = {Open- and Closed-Loop Neural Network Verification Using Polynomial Zonotopes},
  year      = {2023}
}

@inproceedings{ladner_althoff_2023,
  author    = {Ladner, Tobias and Althoff, Matthias},
  booktitle = {Proc. of the Int. Conf. on Hybrid Systems: Computation and Control (HSCC)},
  pages     = {1--13},
  title     = {Automatic Abstraction Refinement in Neural Network Verification Using Sensitivity Analysis},
  year      = {2023}
}

@article{demarchi2024never2,
  title={{NeVer2}: Learning and Verification of Neural Networks},
  author={Demarchi, Stefano and Guidotti, Dario and Pulina, Luca and Tacchella, Armando},
  journal={Soft Computing},
  year={2024}
}

@inproceedings{guidotti2021pynever,
  title={{pyNeVer}: A framework for learning and verification of neural networks},
  author={Guidotti, Dario and Pulina, Luca and Tacchella, Armando},
  booktitle={Automated Technology for Verification and Analysis: 19th International Symposium, ATVA 2021, Gold Coast, QLD, Australia, October 18--22, 2021, Proceedings 19},
  pages={357--363},
  year={2021},
  organization={Springer}
}

@inproceedings{tran2020cav_tool,
author = "Hoang-Dung Tran and Xiaodong Yang and Diego Manzanas Lopez and Patrick Musau and Luan Viet Nguyen and Weiming \
Xiang and Stanley Bak and Taylor T. Johnson",
title = "{NNV}: The Neural Network Verification Tool for Deep Neural Networks and Learning-Enabled Cyber-Physical Systems",
booktitle = "32nd International Conference on Computer-Aided Verification (CAV)",
year = "2020",
month = "July",
}

@inproceedings{manzanas2023cav,
author = "Diego Manzanas Lopez and Sung Woo Choi and Hoang-Dung Tran and Taylor T. Johnson",
title = "{NNV 2.0}: The Neural Network Verification Tool",
booktitle = "35th International Conference on Computer-Aided Verification (CAV)",
year = "2023",
month = "July"
}

@article{manino2025floating,
  title={Floating-Point Neural Network Verification at the Software Level},
  author={Manino, Edoardo and Farias, Bruno and Menezes, Rafael S{\'a} and Shmarov, Fedor and Cordeiro, Lucas C},
  journal={arXiv preprint arXiv:2510.23389},
  year={2025}
}

@article{wilkinson1960error,
  title={Error analysis of floating-point computation},
  author={Wilkinson, James H},
  journal={Numerische Mathematik},
  volume={2},
  number={1},
  pages={319--340},
  year={1960},
  publisher={Springer}
}

@book{demmel1997applied,
  title={Applied numerical linear algebra},
  author={Demmel, James W},
  year={1997},
  publisher={SIAM}
}

@article{hu2026lipnext,
  title={LipNeXt: Scaling up Lipschitz-based Certified Robustness to Billion-parameter Models},
  author={Hu, Kai and Hu, Haoqi and Fredrikson, Matt},
  journal={arXiv preprint arXiv:2601.18513},
  year={2026}
}

@article{beuzeville2026deterministic,
  title={Deterministic and probabilistic rounding error analysis of neural networks in floating-point arithmetic},
  author={Beuzeville, Th{\'e}o and Buttari, Alfredo and Gratton, Serge and Mary, Theo},
  journal={IMA Journal of Numerical Analysis},
  pages={draf130},
  year={2026},
  publisher={Oxford University Press}
}

@article{szasz2025no,
  title={No soundness in the real world: On the challenges of the verification of deployed neural networks},
  author={Sz{\'a}sz, Attila and B{\'a}nhelyi, Bal{\'a}zs and Jelasity, M{\'a}rk},
  journal={arXiv preprint arXiv:2506.01054},
  year={2025}
}

@article{blanchard2020mixed,
  title={Mixed precision block fused multiply-add: Error analysis and application to GPU tensor cores},
  author={Blanchard, Pierre and Higham, Nicholas J and Lopez, Florent and Mary, Theo and Pranesh, Srikara},
  journal={SIAM Journal on Scientific Computing},
  volume={42},
  number={3},
  pages={C124--C141},
  year={2020},
  publisher={SIAM}
}

@article{elarar2025mixed,
  title={Mixed precision accumulation for neural network inference guided by componentwise forward error analysis},
  author={Arar, El-Mehdi El and Filip, Silviu-Ioan and Mary, Theo and Riccietti, Elisa},
  journal={IMA Journal of Numerical Analysis},
  doi={10.1093/imanum/drag054},
  year={2026}
}

@inproceedings{zombori2021fooling,
  title={Fooling a complete neural network verifier},
  author={Zombori, D{\'a}niel and B{\'a}nhelyi, Bal{\'a}zs and Csendes, Tibor and Megyeri, Istv{\'a}n and Jelasity, M{\'a}rk},
  booktitle={International Conference on Learning Representations},
  year={2021}
}

@inproceedings{cordeiro2025esop,
  title={Neural network verification is a programming language challenge},
  author={Cordeiro, Lucas C and Daggitt, Matthew L and Girard-Satabin, Julien and Isac, Omri and Johnson, Taylor T and Katz, Guy and Komendantskaya, Ekaterina and Lemesle, Augustin and Manino, Edoardo and {\v{S}}inkarovs, Artjoms and others},
  booktitle={European Symposium on Programming},
  pages={206--235},
  year={2025},
  organization={Springer}
}

@book{Coq,
  title={Interactive theorem proving and program development: Coq’Art: the calculus of inductive constructions},
  author={Bertot, Yves and Cast{\'e}ran, Pierre},
  year={2004},
  publisher={Springer}
}

@article{demmel1984,
  author  = {Demmel, James},
  title   = {Underflow and the Reliability of Numerical Software},
  journal = {SIAM Journal on Scientific and Statistical Computing},
  volume  = {5},
  number  = {4},
  pages   = {887--919},
  year    = {1984},
  doi     = {10.1137/0905062}
}

% ============================================================================
% SUPPLEMENTAL APPENDICES. Per the POPL CFP, appendices must be submitted as
% separate supplemental material, not part of the main PDF. We keep them here in
% one source file (so \cref works and arXiv carries the full version) and split
% them off the built PDF with pdfjam at submission. The \clearpage below marks
% the split boundary: main PDF = pages up to here; supplement = from here on.
% ============================================================================
\clearpage
\appendix
\crefalias{section}{appendix}  % \cref renders these appendix sections as "Appendix N"

\begin{center}
\makeatletter
{\Large Supplemental Appendices for ``\@title''}
\makeatother
\end{center}

\section{Trusted Axioms: Standard Properties of the Spectral Norm}\label{app:axioms}

This appendix lists in full the axioms that \cref{sec:intro} names as the only ones our Rocq
development introduces (the standard-library axioms it additionally relies on are listed at the
end). We axiomatise the matrix spectral norm as follows: we assume the existence of a
function $\|\cdot\|_2 : \mathbb{R}^{m\times n} \to \mathbb{R}$ satisfying the
axioms below. Each is a standard property of the spectral norm. Below $A, B \in \mathbb{R}^{m\times n}$,
$v \in \mathbb{R}^{n}$, and $c \in \mathbb{R}$, and $|A| \le B$ abbreviates the
entrywise domination $|A_{ij}| \le B_{ij}$ for all $i,j$. The Frobenius norm
appearing in the last two axioms is not itself axiomatised. It is
instead defined as $\|A\|_F \defeq \sqrt{\textstyle\sum_{i,j} A_{ij}^{2}}$.
From this definition and the axioms below we derive all other properties that we use.
\begin{align*}
0 &\le \|A\|_2 && \text{(non-negativity)} \\
\|A v\|_2 &\le \|A\|_2\,\|v\|_2 && \text{(consistency with the vector norm)} \\
\|A+B\|_2 &\le \|A\|_2 + \|B\|_2 && \text{(triangle inequality)} \\
\|c A\|_2 &= c\,\|A\|_2 && \text{(absolute homogeneity, } c \ge 0\text{)} \\
\|\mathbf{0}\|_2 &= 0 && \text{(definiteness)} \\
|A| \le B &\Rightarrow \|A\|_2 \le \|B\|_2 && \text{(entrywise monotonicity)} \\
\|A\|_2 &\le \|A\|_F && \text{(dominated by the Frobenius norm)} \\
\|A^\top A\|_2 &= \|A\|_2^{2} && \text{(Gram identity)} \\
\|A^\top\|_2 &= \|A\|_2 && \text{(transpose invariance)}
\end{align*}
Entrywise monotonicity is Higham's Lemma~6.6(b); the remainder are entirely
standard~\cite{higham2002}. Transpose invariance is specific to the spectral
norm---it fails for general matrix norms---and justifies iterating on the smaller
of $M^\top M$ and $M M^\top$ (\cref{sec:norms}). Definiteness is derivable from
absolute homogeneity and is retained only for convenience.
The existence of $\|\cdot\|_2$ and the first six properties are declared in
\texttt{radii.v}; the last three (\texttt{spectral\_le\_frob},
\texttt{spectral\_gram}, \texttt{spectral\_norm\_trmx}) in
\texttt{gram\_iteration.v}, and only the Gram results of \cref{sec:norms}
depend on them.

\begin{sloppypar}
Beyond these ten, the development relies only on axioms of Rocq's standard
library. Five enter through Rocq's classical real numbers and are used by
most results: \texttt{classic} (classical logic);
\texttt{epsilon\_statement} (Hilbert's epsilon operator);
\texttt{functional\_extensionality\_dep}; and \texttt{sig\_forall\_dec} and
\texttt{sig\_not\_dec}, the two decidability axioms of the classical Dedekind
reals. The results of \cref{sec:hybrid} additionally use \texttt{eq\_rect\_eq}
(uniqueness of identity proofs, module \texttt{Eqdep.Eq\_rect\_eq}), which
dependent pattern matching on the layer-indexed records of high-precision
measurements introduces. \texttt{Print Assumptions} on the mechanised
statement of each numbered result lists a
subset of these sixteen names and nothing else.
\end{sloppypar}

\clearpage

\section{The Counterexample Searches}
\label{app:cex-figs}

\begin{figure}[h]
  \centering\includegraphics[width=\textwidth]{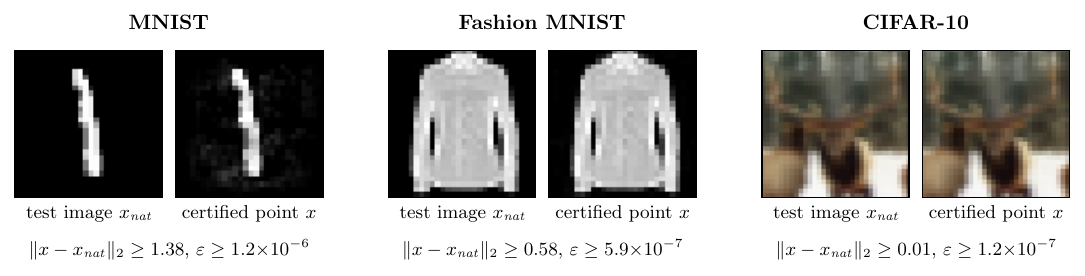}
  \caption{Counterexamples to the certificates of Tobler et al.'s verified certifier, one
    per naturally trained float32 model.\label[suppfigure]{fig:cex_float32} In each pair,
    $\xnat$ (left) is the test image the search started from and $x$ (right) is certified
    robust at $\varepsilon$, yet there exists $x'$ within $\varepsilon$ of $x$ that is
    classified differently under float32 execution. Distances and radii are rounded down.}
\end{figure}

This appendix accompanies \cref{sec:cex}: it details the continuous search that produced the
counterexamples for the naturally trained models at each format and the grid search run from
the test images, how their output is verified, and shows representative counterexamples
(\cref{fig:cex_float32}).

\paragraph{Continuous search}
The search is parametric in the \emph{deployed execution} being attacked: every counterexample
judgement---the classification of each candidate point, and each certification check---is made
against that execution, so the returned pairs are counterexamples for it. The counterexamples
reported in \cref{sec:cex} are against the IEEE-754-compliant execution described there, at
float32 and, for the radii it reports at float16 and float64, at those formats, with the
float32 weights cast to the format (at float16, the deployment rounding of the weights).

The search takes a starting point $\xnat$. From this it finds a nearby point~$\xtie$ that is
classified differently to $\xnat$ by applying the DeepFool~\cite{DeepFool} adversarial search
procedure, as implemented in the Adversarial Robustness Toolbox~\cite{ART} library,
version 1.20.1, run against the TensorFlow model. DeepFool serves only as a heuristic for
locating the decision boundary: if its crossing does not transfer to the deployed execution
(the two executions' boundaries are offset), the search walks further along the ray from
$\xnat$ through $\xtie$ until the deployed execution's classification flips. Linear
interpolation is then applied, with binary search, to find on the line between $\xnat$ and
$\xtie$ an initial pair of points $(\xbinit, x')$ such that $\xbinit$ is certified robust at
$\|\xbinit - x'\|_2$ but $x'$ lies on the other side of the decision boundary (due to
floating-point rounding). This initial pair is then \emph{expanded} by searching outwards from
$\xbinit$ to find the furthest $x$ that is classified the same as $\xbinit$ and certified
robust at $\varepsilon = \|x - x'\|_2$. The search procedure returns the final
$(x, x', \varepsilon)$ that it found.

\paragraph{Verification}
The search's arithmetic is float64, but every point it judges, and every point it returns, is
first cast to the deployed execution's format. Each
returned triple is then re-checked, independently of the search, in exact rational
arithmetic: $\|x' - x\|_2 \le \varepsilon$ (compared as squares, so no square root is taken),
the deployed execution classifies $x$ and $x'$ differently, and the certifier's margin test
holds for $x$'s output under that execution with the Lipschitz constants read exactly from the
output of Tobler et al.'s certifier. All $30$ triples returned for each model and each of the
three formats pass; they are the counterexamples reported in \cref{sec:cex} and validated
against in \cref{sec:eval} (CIFAR-10 at float16 excepted, \cref{sec:eval-setup}).

\paragraph{Examples}
One counterexample for each of the three models is shown in \cref{fig:cex_float32}.

\paragraph{Grid search}
The grid search of \cref{sec:cex} is run against the IEEE-754-compliant float32 execution
described there, on the three float32 models and their adversarially-biased variants. In the
notation of \cref{sec:nn-model}, the adversarially-biased variants set $b_{L-1} = B\,\mathbf{1}$ at
the last hidden layer and $b_L = -W_L\,b_{L-1}$ at the output layer, with every other bias zero
and $B = 10^6$, $3\times10^6$ and $4\times10^6$ for MNIST, Fashion MNIST and CIFAR-10. The
search starts from each test image~$x$ that the certifier certifies at the standard radius and runs a projected-gradient $\ell_2$ attack~\cite{madry2018} inside
$B(x,\varepsilon)$ on the model executed in float64: $200$ steps of size
$\varepsilon/10$, with $5$ restarts. Every iterate is rounded to the 8-bit grid; if rounding
takes it outside the ball, the pixel values furthest from~$x$ are moved back towards~$x$ one
intensity level at a time until it is inside. Each rounded iterate is classified by the float32
execution, and the first classified differently from~$x$ is returned as~$x'$. Each returned
pair is re-checked in exact rational arithmetic: both images lie on the grid,
$\|x' - x\|_2 \le \varepsilon$ (compared as squares), the float32 classifications differ, and
the certifier's margin test holds for $x$'s float32 output at~$\varepsilon$ with the Lipschitz
constants read exactly from the output of Tobler et al.'s certifier. The search is incomplete:
failing to find an~$x'$ says nothing about~$x$. Counterexamples for the adversarially-biased
models appear in the main text (\cref{fig:cex_mnist_biased}).

\clearpage

\section{The Frobenius-Norm Bound and Runtime Checks of \cref{alg:gramfp}}\label{app:derivations}

This appendix defines the Frobenius-norm upper bound $\FrobUB$ that \cref{sec:norms} describes in
words, shows that it never under-estimates, and states the runtime checks that \cref{alg:gramfp}
and $\FrobUB$ perform.

\paragraph{The Frobenius-norm bound}
For a float64 matrix $X$ with $p$ rows and $q$ columns, $\FrobUB$ computes
$\sigma \defeq \fl{\sum_{i,j} X_{ij}^2}$ in float64, a length-$pq$ dot product of the flattened
matrix with itself, and returns
\[
  \FrobUB(X) \;\defeq\; \SqrtUB\biggl(\frac{\sigma + \bar a_{pq}}{1 - \bar\gamma_{pq}}\biggr),
  \qquad \bar\gamma_n \defeq \frac{nu}{1-nu}, \qquad \bar a_n \defeq (1 + \bar\gamma_{n-1})\,n\,\amul .
\]
By the dot-product forward-error bound behind \cref{lem:gram-fp},
$\|X\|_F^2 - \sigma \le \gamma_{pq}\,\|X\|_F^2 + \adotfwd{pq}$, so
$\|X\|_F^2 \le (\sigma + \adotfwd{pq})/(1 - \gamma_{pq})$; since $\bar\gamma_n \ge
\gamma_n$~\cite[Lemma~3.1]{higham2002} and hence $\bar a_n \ge \adotfwd{n}$, this gives
$\FrobUB(X) \ge \|X\|_F$. The closed forms $\bar\gamma_n$ and $\bar a_n$ stand in for
$\gamma_n = (1+u)^n - 1$, whose exact rational value is prohibitively expensive to materialise
for $n = pq$ in the millions.

\paragraph{The runtime checks of \cref{alg:gramfp}}
Both float64 steps that feed a rounding bound---the Gram product of \cref{lem:gram-fp} and
$\FrobUB$'s sum of squares---require that their float64 result not overflow. $\FrobUB$ and
$\mathrm{GramIterFP}$ therefore each check at runtime that their float64 result is finite, and
$\FrobUB$ additionally that $2pqu < 1$ for its $pq$-term sum, so that $\bar\gamma_{pq} < 1$ and
its closed form is valid (at float64, $pq < 2^{52}$); either check failing refuses to certify.
The rescaling factor $c_k$ is positive by construction ($\FrobUB$'s absolute term is), and only
its positivity matters: it is multiplied back exactly in the backward pass. In practice the
checks never fire, because rescaling keeps every iterate's entries $O(1)$, but we retain them
for soundness.

\section{Models, Training and Evaluation Details}
\label{app:training}

This appendix supplies the details of the experimental setup of \cref{sec:eval-setup} that the
body omits: the models and their training, the Gram-iteration counts, and the settings of the
attacks that \cref{sec:rq2} runs on CIFAR-10.

The three image classifiers (MNIST, Fashion MNIST, CIFAR-10) are those of Tobler et
al.~\cite{tobler2025}; we refer the reader to their paper for the corresponding training
details. Writing $[128]{\times}5$ for five hidden layers of 128 neurons and $\cat$ for the
concatenation of such runs, their architectures are $[128]{\times}8$ (MNIST),
$[256] \cat [128]{\times}11$ (Fashion MNIST) and $[512,256] \cat [128]{\times}6$ (CIFAR-10), with
inputs of dimension 784 (MNIST, Fashion MNIST) or 3072 (CIFAR-10) and 10 output classes. The HIGGS and EMNIST classifiers are ours, trained with globally-robust (gloro)
training~\cite{leino2021} using the \texttt{gloro} library (TensorFlow). \Cref{tab:training} gives the
per-model configuration. All of these models share the following settings: the Adam
optimiser with initial learning rate $10^{-3}$ decayed to $10^{-6}$ over training; a fixed
training perturbation radius (no $\varepsilon$ schedule); the gloro sparse-categorical
cross-entropy loss with no TRADES term; and no data augmentation. Weights and activations
are float32.

\begin{table}[h]
\centering
\small
\begin{tabular}{l l c c c r r}
\toprule
Model & Architecture & \#cls & $\varepsilon_{\mathrm{train}}$ & $\varepsilon_{\mathrm{cert}}$ & Epochs & Batch \\
\midrule
HIGGS-128       & $[128]{\times}5$         & 2  & 0.1 & 0.1 & 15  & 512 \\
HIGGS-256       & $[256]{\times}5$         & 2  & 0.1 & 0.1 & 15  & 512 \\
HIGGS-512       & $[512]{\times}5$         & 2  & 0.1 & 0.1 & 15  & 512 \\
HIGGS-1024      & $[1024]{\times}5$        & 2  & 0.1 & 0.1 & 15  & 512 \\
EMNIST-ByClass  & $[512,256] \cat [128]{\times}6$ & 62 & 0.4 & 0.3 & 300 & 256 \\
EMNIST-Balanced & $[512]{\times}8$         & 47 & 0.4 & 0.3 & 500 & 256 \\
\bottomrule
\end{tabular}
\caption{\label{tab:training}%
  Training configuration for the models we trained ourselves (the image models are Tobler
  et al.'s~\cite{tobler2025}). \emph{Architecture} lists hidden-layer widths (all
  fully-connected, ReLU; $[128]{\times}5$ is five layers of 128, and $\cat$ concatenates such runs);
  \emph{\#cls} is the
  number of classes; $\varepsilon_{\mathrm{train}}$ and $\varepsilon_{\mathrm{cert}}$ are the
  training and certification radii; \emph{Epochs} is the number of training epochs and
  \emph{Batch} the mini-batch size.}
\end{table}

HIGGS is a tabular binary-classification task with 28 standardised continuous features; we
use the canonical Baldi et al.~\cite{baldi2014} split (the first $10{,}500{,}000$ examples for
training and the final $500{,}000$ for testing), and train and certify at the same radius
$\varepsilon = 0.1$. The features are standardised per-feature to zero mean and unit variance, with
statistics computed on the training split only (no test leakage) and this same transformation
applied to the evaluation set; consequently $\varepsilon = 0.1$ is an $\ell_2$ radius in
standardised (per-feature-$\sigma$) input space. Following the usual gloro convention of training at a slightly larger
radius than is certified, the EMNIST models are trained at $\varepsilon = 0.4$ and certified at
$\varepsilon = 0.3$; EMNIST-ByClass reuses CIFAR-10's $[512,256] \cat [128]{\times}6$ architecture
(\cref{sec:eval}), while EMNIST-Balanced is eight layers of~512.

The \texttt{gloro} library reports a training-time robustness estimate for each model; these
closely match the sound real-arithmetic verified robust accuracies of \cref{tab:master}
(e.g.\ $62.7\%$ vs.\ $62.7\%$ for HIGGS-1024, and $75.4\%$ vs.\ $75.4\%$ for EMNIST-ByClass),
corroborating that our real-arithmetic baseline reproduces the models' intended robustness.

\paragraph{Gram-iteration counts}
Every result uses 12 Gram iterations except \textbf{RQ1}, which checks the counterexamples with
tighter per-model counts: 20 for MNIST, 13 for Fashion MNIST and 12 for CIFAR-10.

\paragraph{CIFAR-10 attack settings}
The attacks of \cref{sec:rq2} run on all $10{,}000$ CIFAR-10 test points, each attacked away
from the label that the float32 numpy execution of \cref{sec:eval-setup} predicts for it:
PGD-$\ell_2$ and, independently, AutoAttack's APGD-CE then APGD-DLR~\cite{croce2020autoattack},
each with $100$ steps and $5$ restarts. The attacks run at radius $\varepsilon(1 - 10^{-5})$ so
that their float32 projection lands inside the $\varepsilon$-ball; an adversarial point refutes
robustness iff its distance from the test point, computed exactly, is at most
$\varepsilon = 0.141$ and the numpy execution's label differs, the judgement of \cref{sec:cex}.
PGD refutes $3{,}148$ points and APGD $3{,}342$, a superset of PGD's; no adversarial point fell
outside the ball.

\section{Flush-to-Zero Execution}\label{app:ftz}

The standard model of \cref{sec:float-error} assumes gradual underflow (subnormal support), the
IEEE-754 default. Some runtimes instead \emph{flush to zero} (FTZ): a result that would be
subnormal is replaced by~$0$, and, under the companion \emph{denormals-are-zero} (DAZ) mode, a
subnormal \emph{operand} is treated as~$0$ before the operation. These modes trade the standard
model's guarantee for speed, and they are not exotic: validating TensorFlow's inference path
against the standard model---by executing single multiplies and additions through it and
comparing against an independently round-to-nearest computed reference---we found that its
float16 executions conform exactly, but at float32 and float64 it enables FTZ and DAZ, with no
supported way to disable them. (Our experimental platform, \cref{sec:eval}, therefore does not
use TensorFlow and we validated empirically that it conforms to the semantics of \cref{sec:float-error}.) This appendix shows that our theory extends to FTZ runtimes by enlarging the
absolute-error constants, that the extension is \emph{input-independent} (it depends only on the
network architecture and the format), and that a certificate proved with the enlarged constants is
sound for gradual-underflow \emph{and} flush-to-zero execution alike, at a cost governed by a single
format constant that is minute at float32 and float64.

\paragraph{Per-operation model}
Write $\lambda$ for the smallest positive \emph{normal} value ($2^{-126}$ at float32, $2^{-1022}$
at float64; the underflow threshold of~\cite{higham2002}). A flush occurs only when the exact
result already has magnitude below~$\lambda$, so the error it introduces is itself below~$\lambda$,
\emph{independent of the operands' magnitudes}. The FTZ counterpart of the basic-operation
model~\cref{eqn:mixed-model-basic-ops} is therefore the same relation with an enlarged absolute
term, together with one structural change---additions, exact under gradual underflow, may now
flush:
\begin{equation}\label{eqn:ftz-model}
  \fl{x \circ y} = (x \circ y)(1+\delta) + \eta, \qquad |\delta| \le u, \quad |\eta| \le \lambda,
  \quad \delta\eta = 0,
\end{equation}
now including $\circ = {+}$ (for which $\eta = 0$ under gradual underflow). This dominates the
standard model of \cref{sec:float-error}: there, a subnormal result is \emph{rounded} rather than
flushed, and since the spacing at the subnormal scale is $2u\lambda$, its absolute error is at most
$\amul \le u\lambda < \lambda$ (a factor $1/u$ smaller: $\amul \approx 2^{-150}$ versus
$\lambda = 2^{-126}$ at float32), while additions incur no absolute error at all. Every bound
derived from \cref{eqn:ftz-model} therefore also holds under gradual underflow, so it is a common
relaxation of both regimes and a certificate proved from it is sound for either.

\begin{table}[h]
\centering
\small
\setlength{\tabcolsep}{5pt}
\begin{tabular}{@{}lll@{}}
\toprule
Quantity & Standard model (gradual underflow) & Flush-to-zero counterpart \\
\midrule
per-op absolute error $|\eta|$ & $\le\amul$ for $\times,\div$;\ \ $0$ for $+$ & $\le\lambda$ for every op \\
\quad absolute roundoff $\amul$ & $\le u\lambda\ \ (\approx 2^{-150})$ & $\lambda\ \ (= 2^{-126})$ \\
dot-product term $\adot{n}$ & $(1+\gamma_n)\,n\,\amul$ & $(1+\gamma_n)\,(2n-1)\,\lambda$ \\
amplification $\alpha_\ell$ & $L(\varphi_\ell)\bigl(\|W_\ell\|_2+\kappa_{n_\ell}\asnorm{W_\ell}\bigr)$ & unchanged \\
fresh error $\beta_\ell(r)$ & $\cdots+(1+u)\,\adot{n_\ell}\sqrt{m_\ell}$ & $\adot{}\!\to\!\adot{}^{\mathrm{FTZ}}$;\ \ $+\,L(\varphi_\ell)\lambda\sqrt{m_\ell}$ \\
final-layer $\beta_L^{(j,i^*)}(r)$ & $\cdots+2(1+u)\,\adot{n_L}$ & $\adot{}\!\to\!\adot{}^{\mathrm{FTZ}}$;\ \ $+\,2\lambda$ \\
deviation base case $D_0$ & $0$ & $\lambda\sqrt{n_1}$ \\
overflow term (\cref{sec:overflow}) & $\adotfwd{n}$ & $\adot{n}^{\mathrm{FTZ}}$ \\
\bottomrule
\end{tabular}
\caption{\label{tab:ftz}%
  The gradual-underflow quantities of \cref{sec:float-error,sec:overflow,sec:deviation,sec:robustness}
  and their flush-to-zero counterparts, derived in the text below;
  $\adot{n}^{\mathrm{FTZ}} \defeq (1+\gamma_n)(2n-1)\lambda$, and numeric values are for float32. Only the
  absolute-error quantities change: $\gamma_n$, $\kappa_n$, $\alpha_\ell$ and every derivation are
  untouched.}
\end{table}

\paragraph{Propagated constants}
\Cref{tab:ftz} summarises the changes derived here.
Only the absolute-error quantities change; the relative factors $\gamma_n$, $\kappa_n$ are
untouched. A length-$n$ dot product performs $n$ multiplies and $n-1$ additions, each now
contributing up to~$\lambda$, so the dot-product absolute term $\adot{n}$ of \cref{sec:float-error}
is replaced by
\[
  \adot{n}^{\mathrm{FTZ}} \;\defeq\; (1+\gamma_n)\,(2n-1)\,\lambda .
\]
Consequently, in the deviation recursion of \cref{sec:deviation}, the amplification factor
$\alpha_\ell$ is unchanged (it is purely multiplicative, and FTZ adds only absolute error), while
the fresh-error term $\beta_\ell$ gains the enlarged dot-product term and a bias-addition flush
term ($m_\ell$ additions, each up to~$\lambda$, contributing $\lambda\sqrt{m_\ell}$ in $\ell_2$):
\[
  \beta_\ell^{\mathrm{FTZ}}(r) \;\defeq\; L(\varphi_\ell)\Bigl(
    \kappa_{n_\ell}\,\asnorm{W_\ell}\,r + u\,\|b_\ell\|_2
    + (1+u)\,\adot{n_\ell}^{\mathrm{FTZ}}\sqrt{m_\ell}
    + \lambda\sqrt{m_\ell}\Bigr).
\]
The final-layer term $\beta_L^{(j,i^*)}$ of \cref{sec:robustness} gains the same way: $\adot{n_L}$
becomes $\adot{n_L}^{\mathrm{FTZ}}$, and the two output-layer bias additions---one for class~$i^*$,
one for~$j$---may each flush, adding~$2\lambda$ to the pairwise margin bound. The overflow
conditions of \cref{sec:overflow} take the same substitution $\adot{}\mapsto\adot{}^{\mathrm{FTZ}}$.
Flushing and DAZ only \emph{shrink} operand and result magnitudes, with one exception: dropping a
subnormal term that was cancelling against the running value can \emph{raise} a magnitude, by at
most~$\lambda$ per operation. Because $\adot{n_\ell}$ therefore no longer bounds the executed
forward error, the substitution is \emph{necessary}, not merely convenient: the gradual-underflow
constant would be unsound here. With $\adot{}^{\mathrm{FTZ}}$---which does bound the accumulated
drift---in its place, the left-hand side is again a true upper bound on the executed magnitude, so
the condition is a sound sufficient condition for overflow-freedom under FTZ. It differs from the
gradual-underflow condition only by a $\lambda$-scale amount, so it is negligibly stricter in
practice, but the guarantee it yields is exact, not approximate. Finally, DAZ can zero a subnormal \emph{network input} component, provided the
certifier checks---as ours can, from the exact parameters---that no weight or bias is subnormal
(each is zero or has magnitude $\ge\lambda$); internal activations, being outputs of flushed
operations, are never subnormal. A subnormal input component perturbs the input by less
than~$\lambda$, so the recursion's base case becomes $D_0 \defeq \lambda\sqrt{n_1}$ in place of~$0$.
This must hold uniformly over $B(x,\varepsilon)$---where, since $\varepsilon \gg \lambda$, some $x'$
can have subnormal components even when~$x$ does not---so it is a base-case term, not a check
dischargeable at the centre~$x$. These are the only changes; the recursion, the certification
condition (\cref{thm:fp-robust}), and their derivations are otherwise verbatim, and every new
quantity is a function of the architecture and~$\lambda$ alone.

\paragraph{Soundness and magnitude}
Because \cref{eqn:ftz-model} dominates the standard model term by term, the degraded certification
condition instantiated with $\adot{}^{\mathrm{FTZ}}$, $\beta^{\mathrm{FTZ}}$, and
$D_0 = \lambda\sqrt{n_1}$ is sound for any round-to-nearest execution, whether or not it flushes---so
a single certificate covers both regimes. Every quantity added above is bounded by a modest
architecture-dependent multiple of~$\lambda$, which is minute at high precision
($\lambda \approx 1.2\times10^{-38}$ at float32, $2^{-1022}$ at float64) and appreciable only at
float16 ($\lambda = 2^{-14}$)---where certification is already vacuous (\cref{sec:eval}).

\paragraph{Mechanisation}
The recursion, certificate, and overflow conditions are mechanised (Rocq) for the gradual-underflow instantiation; the flush-to-zero results of this appendix are not yet mechanised. Doing so faithfully is a genuine undertaking rather than a per-operation substitution: the mechanised overflow and deviation arguments are specialised to the gradual-underflow operation model, so extending them re-casts the development generically over the per-operation error model, with gradual underflow and flush-to-zero as instances.

\section{The GPUPoly Comparator Configuration}\label{app:gpupoly}

This appendix describes the configuration of GPUPoly~\cite{GPUPoly} that \cref{tab:eran} compares
against: what GPUPoly computes and where floating point enters it, what its shipped
configurations certify, the change we make and the choices behind it, every configuration we
ran, and the two patches. All numbers are for ELINA at commit \texttt{f524156}, the version ERAN's
distribution builds.

\paragraph{What GPUPoly computes}
GPUPoly is the GPU implementation of DeepPoly~\cite{DeepPoly}. Every neuron receives a symbolic
lower and upper bound, an affine expression over the neurons of the previous layer, together with
a concrete interval. To bound a neuron concretely the symbolic bounds are substituted layer by
layer back to the input (\emph{back-substitution}) and the resulting affine expression over the
input box is evaluated. In GPUPoly the substitution is a sequence of matrix products between the
coefficient matrix of the expression being substituted and the weight (or relaxation) matrix of
the layer being crossed, and the final evaluation is a matrix--vector product against the input
box. The analysis is templated on a scalar type $T$ and every one of these products is computed
in interval arithmetic over $T$: in double precision when the input box is handed to it as
float64, in single precision when it is handed to it as float32 (the weights may be either). The two are
separate instantiations of the same code, selected by which entry point the ERAN front end calls.

Floating point enters this computation twice, and the two must be kept apart. First, the
analysis's own arithmetic rounds; GPUPoly makes that sound by directed rounding: every interval
operation rounds its lower end down and its upper end up using the CUDA directed-rounding
intrinsics. This is the ordinary soundness of an interval analysis and is independent of what
the certificate claims about the network. Second, the \emph{network's} execution rounds, and a
certificate for that execution must account for it. GPUPoly does so with a \emph{rounding
model}: in its ``strong floating-point soundness'' build (the default, and the one ERAN ships),
each product $p$ that occurs in the back-substitution products and in the final evaluation is
computed to nearest in $T$ and then pushed one unit in the last place of $T$ outward in each
direction, by \texttt{nextafter}, before being added into the directed-rounded sum. Nothing is
charged for the additions; the source comments the intent as ``user evaluates with precision $T$,
in any order for additions''. The rounding model is the only place in which the certified
execution is specified, and because the step it takes is one ulp \emph{of $T$}, the execution it
certifies has the precision the analysis runs in. This is why GPUPoly's shipped configurations
are what they are: by its authors' account, the double-precision instantiation certifies a
float64 execution under any rounding mode and any summation order (its certification
level~3), and the single-precision instantiation a float32 execution under the same conditions
(level~4). ERAN's own
front end hands GPUPoly float64 weights and a float64 input box, so it always obtains level~3;
to obtain level~4 we gave ERAN's front end float32 weights (exact, since the networks are
float32) and a float32 input box rounded inward from the float64 one (so that it lies within it), which is the
\emph{level~4} column of \cref{tab:eran} (the front-end patch at the end of this appendix).

\paragraph{The change}
Neither shipped configuration certifies our object, a round-to-nearest execution in a chosen
format. We keep the double-precision instantiation, and so the double-precision analysis, the
float64 weights and the float64 input box, and change only the step its rounding model takes.
In the four product lines
of the double-precision instantiation, the one-ulp step \texttt{nextafter}$(p,\pm\infty)$ is
replaced by $p \pm (u\,|p| + \nu)$, computed with directed rounding, where $u$ and $\nu$
are compile-time constants: the unit roundoff and the smallest subnormal of the target format
$F$, the format whose round-to-nearest execution is to be certified. For float32 these are $2^{-24}$ and $2^{-149}$,
for float64 $2^{-53}$ and $2^{-1074}$ (we take the full subnormal rather than the half used by
the round-to-nearest model of \cref{sec:float-error}, which is marginally conservative and far
below every coefficient in these networks). The single-precision instantiation is not touched and
not used. No other line of GPUPoly or ERAN changes; the analysis precision, the abstraction, the
input box, and the accounting of \emph{where} rounding is charged (at products, not at
additions) are all as shipped.

\paragraph{Design of the modification}
Two choices in this change call for explanation: keeping the double-precision analysis, and
charging nothing for the additions. The first is forced. The single-precision instantiation
handicaps the analysis itself: with its own arithmetic in float32, the directed rounding of
every interval operation compounds through the back-substitution, and it certifies almost
nothing on four of our five models (the \emph{level~4} column). A comparison against it would measure that handicap, not the cost of accounting for
the network's rounding. The second choice we inherit from GPUPoly, and it changes meaning
under the first. GPUPoly charges the additions nothing, and as shipped that is sound because
the analysis performs the same additions, in the same precision $T$, with outward rounding: a
sum rounded in any mode never falls below the same operands' sum rounded down, and rounding is
monotone, so the execution's $T$-precision partial sums lie within the analysis's
directed-rounded ones (for the analysis's summation order; that the certificate also covers
other orders is GPUPoly's authors' claim, which we take as given). In our configuration the
analysis adds in double and the certified execution adds in float32, so an addition's float32
rounding, up to $2^{-24}$ relative, is covered only to the $2^{-53}$ of the analysis's own
rounding: the float32 additions are under-charged. Charging them would mean widening every
interval addition of the double-precision path as well, a larger change to a construction we
did not write, and it is not needed for our purpose. With products charged exactly the
standard-model bound and additions free, the modified build verifies at least what GPUPoly's
own accounting verifies at float32 with its additions charged, and on every instance of every
evaluation set at least what level~4 verifies; every comparison in \cref{sec:rq3} is drawn
against it, so it is generous to GPUPoly.
We also leave the input box as
shipped, which has the same direction: the rounding of an input to float32 inside the float64
box (half an ulp of the input at the box boundary) is not charged, whereas our certificate
takes the float32 input the deployed network receives. At float64 none of this arises: the
comparator at float64 reproduces the shipped level~3 to the instance (below), so the patch
changes no float64 verdict and bears only on the comparator column.

\paragraph{The configurations}
Every GPUPoly configuration in the paper is the stock tool, or the stock tool with one of two
patches; no configuration combines them.
The stock library is built from ELINA at commit \texttt{f524156} (the GPUPoly that ERAN's
distribution builds) and run behind ERAN's own front end (the \texttt{lort/eran} image at
digest \texttt{c8c60daa}, whose ERAN is commit \texttt{973e3b5}). The \emph{comparator at $F$}
changes the library and not the front end: the rounding-model patch below, built with the
constants of $F$; \emph{comparator} alone means the comparator at float32. \emph{Level~4}
changes the front end and not the library.
\begin{center}
\footnotesize
\setlength{\tabcolsep}{4pt}
\begin{tabular}{l l l l}
\toprule
Configuration & GPUPoly library & ERAN front end & Reported in \\
\midrule
level~3 & stock & stock & \cref{tab:timing} \\
level~4 & stock & front-end patch & \cref{tab:eran} \\
comparator (at float32) & rounding-model patch, $2^{-24}$, $2^{-149}$ & stock & \cref{tab:eran,tab:timing}, \cref{sec:rq3} \\
comparator at float64 & rounding-model patch, $2^{-53}$, $2^{-1074}$ & stock & below \\
comparator at float16 & rounding-model patch, $2^{-11}$, $2^{-24}$ & stock & \cref{sec:eval-setup}, below \\
real-arithmetic ceiling & rounding-model patch, $0$, $0$ & stock & \cref{tab:eran} (\emph{Real}), below \\
\bottomrule
\end{tabular}
\end{center}
Every library, the stock one included, is built with the same fix to ELINA's build files,
two missing \texttt{thrust} includes and the CUDA architecture of the A10, which touches no
arithmetic. Both patches are reproduced verbatim, so this appendix is self-contained.

\paragraph{Relations among the configurations}
The comparator at float64 and the real-arithmetic ceiling give the level-3 verdict on
every instance of every full evaluation set. The first shows that the patch changed nothing but
the widening step, which at double precision is far below every margin; the second that
GPUPoly's shipped rounding model costs nothing at float64 on these networks, so that level~3
coincides with the real-arithmetic ceiling of its abstraction.
The comparator at float16 verifies nothing on any full evaluation set (on three of the five
models the widened point box no longer even classifies the input; on Fashion MNIST it
classifies $4$ and on MNIST $6{,}244$, and verifies none).
Per instance, on every image model, level~4 verifies a subset of what the comparator verifies,
and the comparator a subset of what the real-arithmetic ceiling verifies.

\emph{The rounding-model patch} is the unified diff below, applied with \texttt{patch -p1} to ELINA at commit
\texttt{f524156}. The inserted block defines the widened step; in each of the
four double-precision \texttt{fma} specialisations the downward and upward
\texttt{nextafter} steps become \texttt{GP\_WID\_DN} and \texttt{GP\_WID\_UP} of the same
argument; the single-precision specialisations are unchanged; and a one-line addition to
\texttt{bindings.cu} prints the constants after the version string, so that every result file
records the model that produced it. The constants of the table above are passed as the
compiler flags \texttt{-DGPUPOLY\_U} and \texttt{-DGPUPOLY\_DEN}, in C hexadecimal
(\texttt{0x1p-24} for $2^{-24}$). That the four changed lines are the only place the
double-precision rounding model is applied is a reading of the source; the patch is the whole
change, so that the reading can be audited against it.
\begin{lstlisting}[language=diff, basicstyle=\scriptsize\ttfamily, breaklines=true,
  breakindent=1.5em, postbreak=\mbox{\textcolor{gray}{$\hookrightarrow$}\space},
  columns=fullflexible, keepspaces=true, mathescape=false, tabsize=1]
--- ELINA/gpupoly/intv.h	2026-10-15 11:33:02.486285134 +0000
+++ ELINA_model/gpupoly/intv.h	2026-10-15 14:52:32.900754705 +0000
@@ -168,12 +168,24 @@
 template <> template<> inline __device__ void convert< Intv<float>>::operator()(Intv<float>& dest, const double& source) { dest = Intv<float>(__double2float_rd(source), __double2float_ru(source)); }
 */
 
+// GPUPOLY-MODEL patch (float-conservatism, 2026-10-15): per-product rounding model of the double-precision analyser.
+// Stock: nextafter, one ulp(double) outward. With GPUPOLY_U defined: widen the round-to-nearest product p by
+// GPUPOLY_U*|p| + GPUPOLY_DEN (Mine's model, unit roundoff GPUPOLY_U); the analyser's arithmetic stays double.
+#ifdef GPUPOLY_U
+__device__ __forceinline__ double gp_wid_up(double p) { return __dadd_ru(__dadd_ru(p, __dmul_ru(fabs(p), (double)(GPUPOLY_U))), (double)(GPUPOLY_DEN)); }
+__device__ __forceinline__ double gp_wid_dn(double p) { return __dadd_rd(__dadd_rd(p, -__dmul_ru(fabs(p), (double)(GPUPOLY_U))), -(double)(GPUPOLY_DEN)); }
+#define GP_WID_UP(p) gp_wid_up(p)
+#define GP_WID_DN(p) gp_wid_dn(p)
+#else
+#define GP_WID_UP(p) nextafter((p), CUDART_INF)
+#define GP_WID_DN(p) nextafter((p), -CUDART_INF)
+#endif
 #ifdef STRONG_FP_SOUNDNESS
 // strong floating point soundness (i.e. user evaluates with precision T, in any order for additions)
-template<> template<> inline __device__ void Intv<double>::fma(Intv<double>& d, const Intv<double>& a, const Intv<double>& b, const Intv<double>& c) { double prod1 = a.low * b.low;	double prod2 = a.low * b.high;	double prod3 = a.high * b.low;	double prod4 = a.high * b.high;	d.low = __dadd_rd(nextafter(fmin(fmin(prod1, prod2), fmin(prod3, prod4)), -CUDART_INF), c.low);	d.high = __dadd_ru(nextafter(fmax(fmax(prod1, prod2), fmax(prod3, prod4)), CUDART_INF), c.high); }
-template<> template<> inline __device__ void Intv<double>::fma(Intv<double>& d, const Intv<double>& a, const double& b, const Intv<double>& c) { double prod1 = a.low * b;	double prod2 = a.high * b;	d.low = __dadd_rd(nextafter(fmin(prod1, prod2), -CUDART_INF), c.low);	d.high = __dadd_ru(nextafter(fmax(prod1, prod2), CUDART_INF), c.high); }
-template<> template<> inline __device__ void Intv<double>::fma(Intv<double>& d, const double& a, const Intv<double>& b, const Intv<double>& c) { double prod1 = a * b.low;	double prod2 = a * b.high;	d.low = __dadd_rd(nextafter(fmin(prod1, prod2), -CUDART_INF), c.low);	d.high = __dadd_ru(nextafter(fmax(prod1, prod2), CUDART_INF), c.high); }
-template<> template<> inline __device__ void Intv<double>::fma(Intv<double>& d, const double& a, const double& b, const Intv<double>& c) { double prod = a * b; d.low = __dadd_rd(nextafter(prod, -CUDART_INF), c.low);	d.high = __dadd_ru(nextafter(prod, CUDART_INF), c.high); }
+template<> template<> inline __device__ void Intv<double>::fma(Intv<double>& d, const Intv<double>& a, const Intv<double>& b, const Intv<double>& c) { double prod1 = a.low * b.low;	double prod2 = a.low * b.high;	double prod3 = a.high * b.low;	double prod4 = a.high * b.high;	d.low = __dadd_rd(GP_WID_DN(fmin(fmin(prod1, prod2), fmin(prod3, prod4))), c.low);	d.high = __dadd_ru(GP_WID_UP(fmax(fmax(prod1, prod2), fmax(prod3, prod4))), c.high); }
+template<> template<> inline __device__ void Intv<double>::fma(Intv<double>& d, const Intv<double>& a, const double& b, const Intv<double>& c) { double prod1 = a.low * b;	double prod2 = a.high * b;	d.low = __dadd_rd(GP_WID_DN(fmin(prod1, prod2)), c.low);	d.high = __dadd_ru(GP_WID_UP(fmax(prod1, prod2)), c.high); }
+template<> template<> inline __device__ void Intv<double>::fma(Intv<double>& d, const double& a, const Intv<double>& b, const Intv<double>& c) { double prod1 = a * b.low;	double prod2 = a * b.high;	d.low = __dadd_rd(GP_WID_DN(fmin(prod1, prod2)), c.low);	d.high = __dadd_ru(GP_WID_UP(fmax(prod1, prod2)), c.high); }
+template<> template<> inline __device__ void Intv<double>::fma(Intv<double>& d, const double& a, const double& b, const Intv<double>& c) { double prod = a * b; d.low = __dadd_rd(GP_WID_DN(prod), c.low);	d.high = __dadd_ru(GP_WID_UP(prod), c.high); }
 template<> template<> inline __device__ void Intv<float>::fma(Intv<float>& d, const Intv<float>& a, const Intv<float>& b, const Intv<float>& c) { float prod1 = a.low * b.low;	float prod2 = a.low * b.high;	float prod3 = a.high * b.low;	float prod4 = a.high * b.high;	d.low = __fadd_rd(nextafterf(fminf(fminf(prod1, prod2), fminf(prod3, prod4)), -CUDART_INF_F), c.low);	d.high = __fadd_ru(nextafterf(fmaxf(fmaxf(prod1, prod2), fmaxf(prod3, prod4)), CUDART_INF_F), c.high); }
 template<> template<> inline __device__ void Intv<float>::fma(Intv<float>& d, const Intv<float>& a, const float& b, const Intv<float>& c) { float prod1 = a.low * b;	float prod2 = a.high * b;	d.low = __fadd_rd(nextafterf(fminf(prod1, prod2), -CUDART_INF_F), c.low);	d.high = __fadd_ru(nextafterf(fmaxf(prod1, prod2), CUDART_INF_F), c.high); }
 template<> template<> inline __device__ void Intv<float>::fma(Intv<float>& d, const float& a, const Intv<float>& b, const Intv<float>& c) { float prod1 = a * b.low;	float prod2 = a * b.high;	d.low = __fadd_rd(nextafterf(fminf(prod1, prod2), -CUDART_INF_F), c.low);	d.high = __fadd_ru(nextafterf(fmaxf(prod1, prod2), CUDART_INF_F), c.high); }
--- ELINA/gpupoly/bindings.cu	2026-10-15 11:33:02.476285012 +0000
+++ ELINA_model/gpupoly/bindings.cu	2026-10-15 14:52:32.900754705 +0000
@@ -53,6 +53,11 @@
 		std::cout << "GPUPoly " << VERSION_MAJOR << "." << VERSION_MINOR << "." << VERSION_PATCH;
 #ifdef STRONG_FP_SOUNDNESS
 		std::cout << "S";
+#ifdef GPUPOLY_U
+#define GP_STR2(x) #x
+#define GP_STR(x) GP_STR2(x)
+		std::cout << " model(u=" << GP_STR(GPUPOLY_U) << ",den=" << GP_STR(GPUPOLY_DEN) << ")";
+#endif
 #else
 		std::cout << "W";
 #endif
\end{lstlisting}

\emph{The front-end patch} is the unified diff below, applied to \texttt{optimizer.py} and
\texttt{\_\_main\_\_.py} in that ERAN's \texttt{tf\_verify}. Three casts hand GPUPoly float32
weights and biases (exact, since the networks are float32), on which GPUPoly's Python
interface selects its single-precision entry points. The helper rounds the centre to the
float32 input the deployed network receives and rounds the float64 box inward to float32
(lower bounds up, upper bounds down), so that the float32 box lies within the float64 one and
the containment of \cref{sec:eval-setup} covers it; the point box of the correctness test,
already float32, is unchanged. The assertions and the banner, which every result file
records, are part of the change as run.
\begin{lstlisting}[language=diff, basicstyle=\scriptsize\ttfamily, breaklines=true,
  breakindent=1.5em, postbreak=\mbox{\textcolor{gray}{$\hookrightarrow$}\space},
  columns=fullflexible, keepspaces=true, mathescape=false, tabsize=1]
--- eran_orig/optimizer.py	2026-11-05 02:31:11.006686735 +0000
+++ eran_patch/optimizer.py	2026-11-05 02:31:11.489695503 +0000
@@ -501,8 +501,8 @@
                 
                 matrix, bias, m_input_names, b_output_name, b_output_shape = self.resources[i][domain]
                 #print("type ", type(matrix), type(bias), matrix.dtype, bias.dtype)
-                network.add_linear(matrix.astype("float64"))
-                network.add_bias(bias.astype("float64"))
+                network.add_linear(matrix.astype("float32"))  # GPUPOLY-F32
+                network.add_bias(bias.astype("float32"))  # GPUPOLY-F32
                 nn.weights.append(matrix)
                 nn.biases.append(bias)
                 nn.layertypes.append('FC')
@@ -560,7 +560,7 @@
                 network.add_conv_2d(image_shape[0], image_shape[1], filters.astype("float64"), strides[0], [pad_top, pad_left, pad_bottom, pad_right])
                 bias=bias.repeat(b_output_shape[1]*b_output_shape[2])
                 #print("Filter Bias ", bias)
-                network.add_bias(bias.astype("float64"))
+                network.add_bias(bias.astype("float32"))  # GPUPOLY-F32
                 num_gpu_layers +=2
                 i += 1    
            
--- eran_orig/__main__.py	2026-11-05 02:31:11.352693016 +0000
+++ eran_patch/__main__.py	2026-11-05 02:31:11.489695503 +0000
@@ -106,6 +106,26 @@
     print('==================================================================')
 
 
+# GPUPOLY-F32: single-precision input boxes, rounded inward (see patch_eran_float32.py)
+_GPUPOLY_F32_ANNOUNCED = False
+def _gpupoly_f32_box(lb, ub):
+    global _GPUPOLY_F32_ANNOUNCED
+    lb = np.asarray(lb, dtype=np.float64)
+    ub = np.asarray(ub, dtype=np.float64)
+    lb32 = lb.astype(np.float32)
+    ub32 = ub.astype(np.float32)
+    lb32 = np.where(lb32.astype(np.float64) < lb, np.nextafter(lb32, np.float32(np.inf)), lb32).astype(np.float32)
+    ub32 = np.where(ub32.astype(np.float64) > ub, np.nextafter(ub32, np.float32(-np.inf)), ub32).astype(np.float32)
+    assert (lb32.astype(np.float64) >= lb).all() and (ub32.astype(np.float64) <= ub).all()
+    assert (lb32 <= ub32).all()
+    assert lb32.dtype == np.float32 and ub32.dtype == np.float32
+    if not _GPUPOLY_F32_ANNOUNCED:
+        print("GPUPOLY-F32: float32 weights and float32 input boxes (inward-rounded); "
+              "GPUPoly single-precision entry points (test_s) -> certification level 4", flush=True)
+        _GPUPOLY_F32_ANNOUNCED = True
+    return lb32, ub32
+
+
 def normalize(image, means, stds, dataset):
     # normalization taken out of the network
     if len(means) == len(image):
@@ -1366,6 +1386,7 @@
         if config.num_tests is not None and i >= config.from_test + config.num_tests:
             break
         image= np.float64(test[1:len(test)])/np.float64(255)
+        image = image.astype(np.float32).astype(np.float64)  # GPUPOLY-F32: centre = float32(k/255)
         specLB = np.copy(image)
         specUB = np.copy(image)
         if config.quant_step:
@@ -1387,6 +1408,7 @@
             #specLB = np.reshape(specLB, (32,32,3))#np.ascontiguousarray(specLB, dtype=np.double)
             #specUB = np.reshape(specUB, (32,32,3))
             #print("specLB ", specLB)
+            specLB, specUB = _gpupoly_f32_box(specLB, specUB)  # GPUPOLY-F32
             is_correctly_classified = network.test(specLB, specUB, int(test[0]), True)
         else:
             label,nn,nlb,nub,_,_ = eran.analyze_box(specLB, specUB, init_domain(domain), config.timeout_lp, config.timeout_milp, config.use_default_heuristic)
@@ -1426,6 +1448,7 @@
                 prop = int(target[i])
 
             if domain == 'gpupoly' or domain =='refinegpupoly':
+                specLB, specUB = _gpupoly_f32_box(specLB, specUB)  # GPUPOLY-F32
                 is_verified = network.test(specLB, specUB, int(test[0]))
                 #print("res ", res)
                 if is_verified:
\end{lstlisting}

\end{document}